\pdfoutput=1

\documentclass[11pt]{article}

\usepackage[final]{acl} 

\usepackage{times}
\usepackage{latexsym}

\usepackage{blindtext}
\usepackage{hyperref}
\usepackage{enumitem}

\newcommand{\citelink}[2]{\hyperlink{cite.#1}{#2}}

\usepackage[T1]{fontenc}
\usepackage{spverbatim}

\usepackage[utf8]{inputenc}
\usepackage{multirow}

\usepackage{booktabs}
\usepackage{adjustbox}
\usepackage{makecell}
\usepackage{rotating}
\usepackage{xcolor}

\usepackage{graphicx}
\usepackage{subcaption}
\usepackage{svg}

\usepackage{microtype}

\usepackage{inconsolata}

\usepackage{pifont}
\newcommand{\cmark}{\textcolor{green}{\ding{51}}} 
\newcommand{\xmark}{\textcolor{red}{\ding{55}}} 

\title{Political Leaning and Politicalness Classification of Texts}

\author{Matous Volf $^\spadesuit$, Jakub Simko $^\dagger$ \\
  $^\spadesuit$ DELTA – High school of computer science and economics, Pardubice, Czechia \\
  $^\dagger$ Kempelen Institute of Intelligent Technologies, Bratislava, Slovakia \\
  me@matousvolf.cz, jakub.simko@kinit.sk}

\begin{document}

\maketitle

\begin{abstract}
This paper addresses the challenge of automatically classifying text according to political leaning and politicalness using transformer models. We compose a comprehensive overview of existing datasets and models for these tasks, finding that current approaches create siloed solutions that perform poorly on out-of-distribution texts. To address this limitation, we compile a diverse dataset by combining 12 datasets for political leaning classification and creating a new dataset for politicalness by extending 18 existing datasets with the appropriate label. Through extensive benchmarking with leave-one-in and leave-one-out methodologies, we evaluate the performance of existing models and train new ones with enhanced generalization capabilities.
\end{abstract}

\section{Introduction}
Automatic classification of political leaning has many practical uses and has been researched as well. The research has been focused mostly on internet news articles \cite{baly-etal-2020-detect, demidov-2023-political} and posts from social networks \cite{preotiuc-pietro-etal-2017-beyond, xiao-etal-2020-timme}. In contrast, we combine several (12) datasets with different types of text to maximize our model's ability to generalize and to make it applicable to more than a single domain.

Such machine classification can be utilized by aggregator sites that focus on presenting news stories from all ideological perspectives (e.g. AllSides\footnote{\href{https://www.allsides.com}{www.allsides.com}} or Ground News\footnote{\href{https://ground.news}{ground.news}}). As they point out, bias is not necessarily wrong -- \textit{hidden} bias is what misleads and polarizes society. They mitigate it by providing readers with a balanced mix, allowing them to see the full picture and to consider even the opposing arguments. That prevents confirmation bias and the creation of filter bubbles (inside which users are isolated and limited to the content curated by recommendation systems) and echo chambers (where users are repeatedly exposed to the same viewpoints, reinforcing instead of challenging their existing beliefs). These sites could even use the model to identify, as Ground News calls them, blindspots\footnote{\href{https://ground.news/blindspot/about}{ground.news/blindspot/about}} -- news stories that have political undertones and are disproportionately covered by media sources on one side of the political spectrum. Apart from media, AllSides also tracks the political leaning of the top fact checking sources\footnote{\href{https://www.allsides.com/media-bias/fact-check-bias-chart}{www.allsides.com/media-bias/fact-check-bias-chart}}.

More possible use cases include, for example, tools for auditing content published by policy makers to verify their stance. Similar tools could also be employed to reveal hidden political agendas.

Placing texts on the political spectrum is something human reviewers can do by hand. However, this quickly becomes expensive on the large and growing amounts of available data, which has resulted in the need to automate this process.

\textit{Political leaning} is in the context of this analysis categorized into three classes: left, center and right. This is a simplification of the real problem -- both the left and right are actually spectrums by themselves and encompass views ranging from moderate to more extreme positions. A single dimension is a limitation as well and political science recognizes multi-dimensional political leaning systems (classifying stances individually for, e.g., economic policies, human rights recognition, authoritarianism). Moreover, a single text can in fact show support for both sides (for example on different discussed topics), potentially falling into a ``mixed'' class.

With this task, a prerequisite problem arises: if the input text is not about politics, then the behavior and output of the model is undefined. The solution is to filter the input texts based on the so-called \textit{politicalness}. We define politicalness as a binary class: A text is political if its main topic is politics. Otherwise, it is non-political. An ideal definition would take into account not the mere topic, but rather whether the text expresses a political \textit{opinion}. The reason is that some texts (e.g. Wikipedia articles) may be about political topics, but do not in fact contain any signs of supporting a certain political leaning. This kind of classification is, however, much more difficult and requires very fine-grained data, and so we have decided to proceed with the simpler definition.

The current state of research is not satisfactory: As we are going to show, existing works on political leaning tend to create siloed solutions that break on unseen distributions. On these distributions, no rigorous testing has been done and our \textbf{hypothesis is that models trained on one type of text do not transfer to other ones without decrease in accuracy}. Also, while a few existing classification models have been published, no sufficiently large datasets are available for the task of politicalness.

The contributions of this work are the following: We have 
\begin{enumerate}
    \item composed a comprehensive overview of the existing datasets and models for the tasks of classifying politicalness and political leaning,
    \item created a new dataset for classifying politicalness by compiling several existing datasets and extending them with the label,
    \item identified and unified the available data with the political leaning label,
    \item measured the performance of the existing classification models,
    \item trained and tested a set of own models.
\end{enumerate}

Alongside this paper, we publish the complete source code\footnote{Available at \href{https://github.com/matous-volf/political-leaning-prediction}{github.com/matous-volf/political-leaning-prediction}} of our research, including complete results of all our measurements. We made the model available at Hugging Face\footnote{Available at \href{https://huggingface.co/matous-volf/political-leaning-politics}{https://huggingface.co/matous-volf/political-leaning-politics}}.

\section{Related work}

\subsection{Existing political classification works}

While there are numerous works focusing on predicting political leaning, they do not address the issue of filtering non-political texts. We were not able to find any papers dedicated to politicalness by itself either.

Natural language processing (NLP) classification is not the only way of predicting one's political leaning -- for instance, it can also be based on the user and post relationships on a social network. This is what \cite{wing-etal-2016-quantifying} have done by analyzing retweet behavior and audience overlap to infer political scores through a convex optimization framework. \cite{yupeng-etal-2017-ideology-detection} have leveraged heterogeneous link types (follow, mention, and retweet) with a unified probabilistic model to estimate numerical ideological positions. \cite{stefanov-etal-2020-predicting} have even combined this kind of approach with NLP by first applying unsupervised clustering of the users based on the retweets and labeling the clusters with leaning to then train FastText on the texts with those labels.

The NLP task is relatively challenging. Existing works have experimented with employing traditional machine learning techniques -- like regression \cite{preotiuc-pietro-etal-2017-beyond} or SVMs \cite{zhitomirsky-geffet-etal-2016-utilizing} -- as well as deep learning \cite{iyyer-etal-2014-political, baly-etal-2020-detect}. In the case of complex text classification tasks, transformer based models are a good fit. \cite{demidov-2023-political} showed they are suitable for the particular task of political leaning. All the existing models collected and benchmarked in this study are based either on BERT, RoBERTa, DeBERTa or DistilBERT.

In recent years, the capabilities of large language models have advanced greatly, creating new opportunities in various domains, including the task of classifying text by political characteristics. \cite{heseltine-clemm-von-hohenberg-2024-large-language} discuss the option of using LLMs instead of human manual annotation and \cite{tornberg-2024-large-language} even suggests that they are able to outperform crowd workers and experts. \cite{halterman-2025-synthetically} proposes generating synthetic data with generative LLMs and, in fact, two of the datasets used in this research \cite{jones-2024-political-bias, nayak-2024-political-ideologies} have already been created in this exact way. While these huge models may have a very good understanding of political issues and may perform well in zero- or few-shot classification, employing them comes with some downsides, as we discuss in more detail later (section \ref{sec:discussion}).

A survey study \cite{doan-gulla-2022-survey, nemeth-2022-review} and a scoping review \cite{nemeth2023scoping} have been published that provide an overview of the current techniques, approaches and datasets for these tasks. We have found very useful the guidance provided by \cite{burnham-2024-guide}, despite it being focused on stance detection, which is a separate NLP task.

A related, but distinct, task to those researched in this paper is the detection of \textit{political bias} \cite{gangula-etal-2019-detecting} -- whether and to what extent the author expresses a preference for or against a particular political subject. Another close problem is the prediction of the author's political party \cite{hoyland-etal-2014-predicting}.

\subsection{Existing datasets}

All the datasets we have gathered are publicly available to download from the internet. We have searched the Hugging Face Hub\footnote{\href{https://huggingface.co/datasets}{huggingface.co/datasets}}, Kaggle\footnote{\href{https://www.kaggle.com/datasets}{www.kaggle.com/datasets}} and Zenodo\footnote{\href{https://zenodo.org}{zenodo.org}} using a simple and universal query ``politic'', inspected all the search results and collected any relevant items. All the texts are in English and in the context of the United States. Tables \ref{fig:datasets_leaning} and \ref{fig:datasets_politicalness} list all the collected datasets. We have shortened some of the names for the convenience of working with them. In later sections, we analyze the dataset contents in more detail.

\subsubsection{Political leaning datasets}

There is a total of 12 datasets with political leaning labels we have retrieved: Article bias prediction \cite{baly-etal-2020-detect}; BIGNEWSBLN \cite{liu-etal-2022-politics}; CommonCrawl news articles \cite{spliethover-etal-2022-word}; Dem., rep. party platform topics \cite{wolbrecht-2023-democratic, wolbrecht-2023-republican}; GPT-4 political bias \cite{jones-2024-political-bias}; GPT-4 political ideologies \cite{nayak-2024-political-ideologies}; Media political stance \cite{espana-bonet-2023-multilingual}; Political podcasts \cite{naryana-2024-political-podcasts}; Political tweets \cite{van-steyn-2023-political-tweets}; Qbias \cite{haak-schaer-2023-qbias}; Webis bias flipper 18 \cite{chen-etal-2018-learning}; Webis news bias 20 \cite{chen-etal-2020-analyzing}.

\subsubsection{Politicalness datasets}

For politicalness, we have retrieved 18 datasets. 2 of them do explicitly have the desired label: PoliBERTweet \cite{kawintiranon-singh-2022-polibertweet} and Poltical or not \cite{burnham-etal-2024-political}. The rest we have collected by browsing the repositories mentioned above, searching for any textual data of good quality and a vivid range of topics. These we had to annotate ourselves (as described in the methodology section \ref{sec:methodology_data_preprocessing_politicalness}): Amazon reviews 2023 \cite{hou-etal-2024-bridging}; Dialogsum \cite{chen-etal-2021-dialogsum}; Free news \cite{webhose-free-news}; Goodreads book genres \cite{szemraj-2023-goodreads}; IMDB \cite{maas-etal-2011-learning}; IMDB movie genres \cite{heymans-2022-imdb}; Medium post titles \cite{amrrs-2019-medium}; News category \cite{misra-grover-2021-sculpting, misra-2022-news}; PENS \cite{ao-etal-2021-pens}; Recipes \cite{bian-2022-recipe}; Reddit comments \cite{baumgartner-etal-2020-pushshift}; Reddit submissions \cite{baumgartner-etal-2020-pushshift}; Rotten tomatoes \cite{pang-lee-2005-seeing}; Textbooks \cite{open-phi-2023-textbooks}; Tweet topic multi \cite{antypas-etal-2022-twitter}; Yelp review full \cite{zhang-etal-2015-convolutional}.

\subsection{Existing models}

We have searched the Hugging Face Hub\footnote{\href{https://huggingface.co/models}{huggingface.co/models}} for existing models. We have used a simple and universal query ``politic'' and have inspected each of the results to make sure we would collect as many relevant models as possible.

Two of them are unique in that they can be utilized for a wide range of tasks -- they are not trained for and limited to a single one. POLITICS \cite{liu-etal-2022-politics} is a pretrained model on English news articles of politics, produced via continued training on RoBERTa. It cannot be used out-of-the-box on downstream tasks, but should instead be fine-tuned, serving as a replacement of general transformer models, optimized for politics-related texts. Political DEBATE \cite{burnham-etal-2024-political} is an NLI classifier trained for zero-shot and few-shot classification of political documents. These have been tested on both the polticalness and political leaning tasks.

\subsubsection{Political leaning models}

In total, we have found 8 models trained to classify political leaning: BERT political bias fine-tune \cite{velez-2024-bert}; DeBERTa political classification \cite{palmqvist-2024-deberta}; DistilBERT-PoliticalBias \cite{jones-2024-political-bias}; DistilBERT political fine-tune \cite{shrimali-2024-distilbert}; DistilBERT political tweets \cite{newhauser-2022-distilbert} Political bias BERT \cite{bucket-2023-political}; Political bias prediction AllSides DeBERTa \cite{sahitaj-2024-political}; Political ideologies RoBERTa fine-tuned \cite{nayak-2024-roberta}.

\subsubsection{Politicalness models}

We have managed to find 2 models trained to classify politicalness: Classifier main subject politics \cite{gptmurdock-2024-classifier} and Topic politics \cite{silcock-etal-2024-newswire}.

\section{Methodology}

\subsection{Data preprocessing}

All datasets have textual bodies and some additionally contain titles. In these cases, the titles are prepended (with two line breaks inbetween) to the bodies before passing as the input to the model.

Most of the datasets contain some amount of very short texts that do not in fact bear any value to the training or evaluation. For each of these datasets, we have set an appropriate lower bound and dropped all examples that were shorter.

\begin{figure}[t]
    \centering
    \includegraphics[width=\columnwidth]{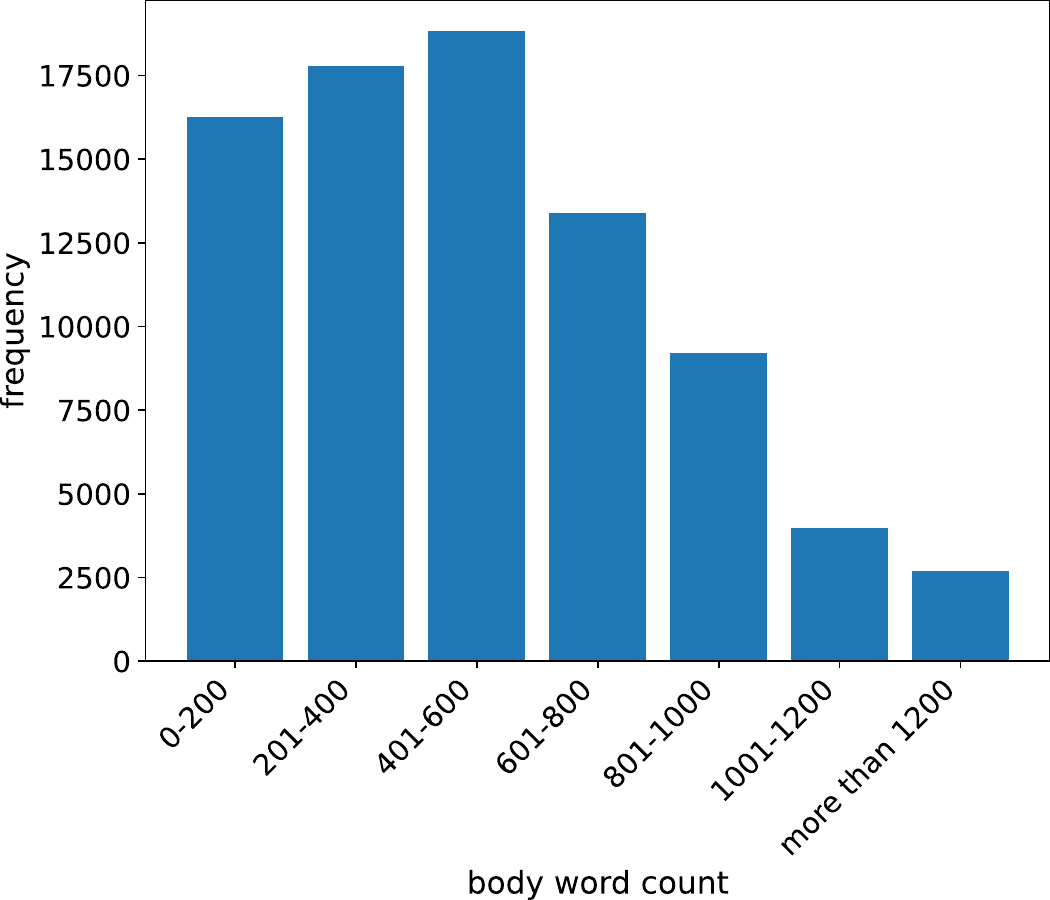}
    \caption{The distribution of the body word count values in the CommonCrawl news articles dataset. The majority of the datasets have a distribution curve very similar to this one, only the maximum text length and the total number of examples are different.}
    \label{fig:dataset_length_distribution}
\end{figure}

Figure \ref{fig:dataset_length_distribution} shows a typical text length distribution that almost all datasets share.

The largest datasets have been downsampled for the purpose of this research, because in the experiments, it is important to maintain balance in the number of examples taken from each dataset. This means that the huge amount of data could never be taken from these large datasets anyway, as that would break this balance in ratio with the small ones. BIGNEWSBLN, CommonCrawl news articles, Media political stance, Political tweets and Yelp review full have been sampled (systematically, per each class, by the text length) to 100~000 datapoints while ensuring an even class distribution.

\subsubsection{Political leaning}\label{sec:methodology_data_preprocessing_leaning}

\begin{figure}[t]
    \centering
    \includegraphics[width=\columnwidth]{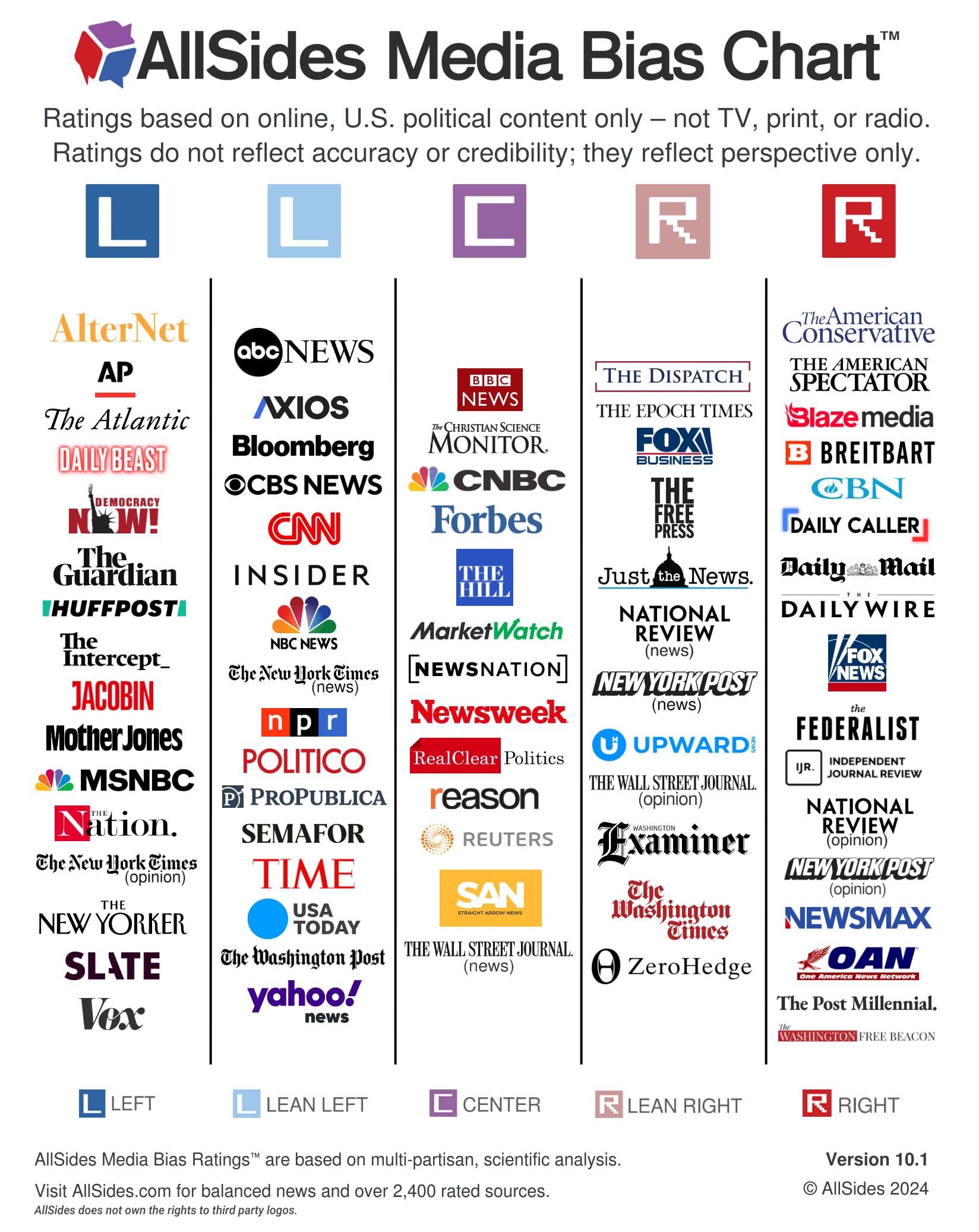}
    \caption{The AllSides media bias chart. Version 10.1, 2024.}
    \label{fig:allsides_media_bias_chart}
\end{figure}

The datasets for political leaning consist of news articles, social network posts, short statements (some synthetically generated by LLMs) and podcast episode descriptions. Table~\ref{fig:datasets_leaning} displays them with some more properties. Some of them do not contain the center leaning label. On the contrary, CommonCrawl news articles and GPT-4 political bias feature 5 classes -- with the left and right sides scaled into two levels (extreme and moderate). The labels have been reduced to 3 classes by mapping both of these levels to the single left or right class, respectively. Since the texts come from the U.S., the left leaning class is mostly defined by the liberal ideology and democratic party views and the same applies for the right class being closely tied to the conservative and republican views. News articles are rated by AllSides. We consider this source of annotations trustworthy because, as noted by \cite{baly-etal-2020-detect}, they are made as a result of a rigorous process that involves blind bias surveys, editorial reviews, third-party analysis, independent reviews and community feedback\footnote{\href{https://www.allsides.com/media-bias/media-bias-rating-methods}{allsides.com/media-bias/media-bias-rating-methods}}. Figure~\ref{fig:allsides_media_bias_chart} shows how the media are categorized at the time of writing of this paper\footnote{\href{https://www.allsides.com/media-bias/media-bias-chart}{allsides.com/media-bias/media-bias-chart}}. The Article bias prediction and Qbias datasets are annotated at the article level, the rest have the labels automatically derived from the leaning rating of the entire media outlet. That brings in some level of inaccuracy, but \cite{baly-etal-2020-detect} have measured that the individually labeled examples differ from the general media leaning in only 3.11~\% of cases, which is acceptable.

After evaluating the existing politicalness models (section~\ref{sec:methodology_existing_evaluation_politicalness}), we have noticed that the best one performs really well (section ~\ref{sec:results_existing_evaluation_politicalness}) on all datasets except for one -- CommonCrawl news articles (which has labels for political leaning, and so is marked entirely as political). On this dataset, it achieved an accuracy of only 33~\% -- lower than a randomly guessing model would. We have investigated the individual examples the model has identified as non-political and confirmed that most of them in fact were. We have considered removing the dataset from the analysis completely, but it is valuable in that it is large and its labels include the center leaning class. For this reason, we have utilized the model as a filter and removed just the non-political texts. To mitigate the removal of false negatives (actually political texts), we have only considered the text non-political if the model was very confident -- with the logit score above 0.99.

\begin{table*}[!ht]
    \small
    \centering
    \begin{tabular}{lrrrrll}
    \toprule
    name & \makecell[r]{total\\count} & \makecell[r]{class (L/C/R)\\distribution} & \makecell[r]{body\\word count\\mean ± std} & \makecell[r]{title\\word count\\mean ± std} & type & \makecell[l]{label\\source} \\ 
    \midrule
    \citelink{liu-etal-2022-politics}{BIGNEWSBLN} & 100~000 & 33 / 33 / 33 & 557 ± \texttt{ } 411 & 9 ± 3 & news & AllSides \\ 
    \citelink{spliethover-etal-2022-word}{CommonCrawl news articles} & 100~000 & 33 / 33 / 33 & 516 ± \texttt{ } 351 & ~ & news & AllSides \\ 
    \citelink{espana-bonet-2023-multilingual}{Media political stance} & 100~000 & 50 / \texttt{ }0 / 50 & 618 ± \texttt{ } 360 & ~ & news & AllSides \\ 
    \citelink{van-steyn-2023-political-tweets}{Political tweets} & 100~000 & 50 / \texttt{ }0 / 50 & 18 ± \texttt{ \ \ } 5 & ~ & social network & author \\ 
    \citelink{baly-etal-2020-detect}{Article bias prediction} & 37~550 & 35 / 29 / 37 & 1~084 ± \texttt{ } 751 & 10 ± 3 & news & AllSides \\ 
    \citelink{haak-schaer-2023-qbias}{Qbias} & 21~715 & 47 / 20 / 33 & 66 ± \texttt{ \ } 28 & 7 ± 3 & news & AllSides \\ 
    \citelink{wolbrecht-2023-democratic}{Dem., rep. party platform topics} & 11~557 & 53 / \texttt{ }0 / 47 & 60 ± \texttt{ \ } 71 & ~ & statements & party \\ 
    \citelink{naryana-2024-political-podcasts}{Political podcasts} & 11~047 & 48 / \texttt{ }0 / 52 & 101 ± \texttt{ \ } 48 & 8 ± 4 & episode blurbs & author \\ 
    \citelink{chen-etal-2020-analyzing}{Webis news bias 20} & 7~722 & 47 / 16 / 37 & 819 ± \texttt{ } 943 & 9 ± 3 & news & AllSides \\ 
    \citelink{chen-etal-2018-learning}{Webis bias flipper 18} & 6~400 & 37 / 24 / 39 & 884 ± 1~128 & 10 ± 3 & news & AllSides \\ 
    \citelink{nayak-2024-political-ideologies}{GPT-4 political ideologies} & 3~200 & 50 / \texttt{ }0 / 50 & 62 ± \texttt{ \ } 10 & ~ & statements & GPT-4 \\ 
    \citelink{jones-2024-political-bias}{GPT-4 political bias} & 612 & 30 / 29 / 41 & 11 ± \texttt{ \ \ } 2 & ~ & statements & GPT-4 \\ 
    \bottomrule
    \end{tabular}
    \caption{The political leaning datasets, ordered by size. The label source: AllSides is described in section \ref{sec:methodology_data_preprocessing_leaning}; Political tweets are written by members of the democratic or republican party; each of the Political podcasts is known to support either liberal or conservative voices; the GPT-4 statements are generated with the label by the LLM.}
    \label{fig:datasets_leaning}
\end{table*}

\subsubsection{Politicalness}\label{sec:methodology_data_preprocessing_politicalness}

Out of the politicalness datasets, two -- PoliBERTweet and Political or not -- have an explicit politicalness label. However, they are not diverse enough (only contain a limited number of topics) and their size is insufficient for training a model. For this reason, we have compiled our own dataset by merging several different datasets (the other 16) containing political and / or non-political texts. They consist of news articles, social network posts and typically non-political texts like user reviews, book and movie summaries or daily dialogues. Table~\ref{fig:datasets_politicalness} displays them with some more properties. When the texts in the dataset were labeled by topics, we have carefully inspected the text contents in each topic in order to either assign the correct label, or discard the topic -- if it was ambiguous -- to mitigate bringing noise to the data.

The Textbooks dataset only consists of 1 795 texts. They are, however, very long -- around 30~000 words on average. We have decided to split them after every 10 paragraphs.

We have also incorporated the political leaning datasets, marking all the examples as political.

The resulting aggregate dataset contains texts of many different types and from diverse sources and does therefore serve the purpose of a universal benchmarking standard. A model trained on it should also be able to generalize well and handle a wide range of possible inputs. To the best of our knowledge, it is the first dataset for politicalness of this size, scope and quality.

\begin{table*}[!ht]
    \centering
    \small
    \begin{tabular}{lrrrrll}
    \toprule
    name & \makecell[r]{non-political\\count} & \makecell[r]{political\\count} & \makecell[r]{body word count\\mean ± std} & \makecell[r]{title word count\\mean ± std} & type & \makecell[l]{label\\source} \\
    \midrule
    \citelink{open-phi-2023-textbooks}{Textbooks} & 262~082 & 1~088 & 202 ± 100 & ~ & other & topics \\ 
    \citelink{baumgartner-etal-2020-pushshift}{Reddit comments} & 207~150 & 0 & 47 ± \texttt{ }79 & ~ & social network & topics \\ 
    \citelink{misra-grover-2021-sculpting, misra-2022-news}{News category} & 94~566 & 24~986 & 24 ± \texttt{ }13 & 9 ± 3 & news & topics \\ 
    \citelink{heymans-2022-imdb}{IMDB movie genres} & 104~063 & 0 & 101 ± \texttt{ }74 & ~ & other & topics \\ 
    \citelink{zhang-etal-2015-convolutional}{Yelp review full} & 100~000 & 0 & 96 ± \texttt{ }60 & ~ & casual & implicit \\ 
    \citelink{maas-etal-2011-learning}{IMDB} & 98~469 & 0 & 233 ± 173 & ~ & casual & implicit \\ 
    \citelink{baumgartner-etal-2020-pushshift}{Reddit submissions} & 82~069 & 0 & 120 ± 179 & 10 ± 6 & social network & topics \\ 
    \citelink{ao-etal-2021-pens}{PENS} & 70~694 & 45~92 & 599 ± 753 & 10 ± 3 & news & topics \\ 
    \citelink{bian-2022-recipe}{Recipes} & 65~390 & 0 & 21 ± \texttt{ \ }9 & ~ & other & implicit \\ 
    \citelink{hou-etal-2024-bridging}{Amazon reviews 2023} & 61~715 & 0 & 69 ± 102 & 4 ± 3 & casual & implicit \\ 
    \citelink{amrrs-2019-medium}{Medium post titles} & 52~171 & 3~567 & 12 ± \texttt{ \ }6 & 8 ± 3 & social network & topics \\ 
    \citelink{antypas-etal-2022-twitter}{Tweet topic multi} & 21~416 & 0 & 28 ± \texttt{ }12 & ~ & social network & human \\ 
    \citelink{kawintiranon-singh-2022-polibertweet}{PoliBERTweet} & 10~000 & 10~000 & 23 ± \texttt{ }15 & ~ & social network & hashtags \\ 
    \citelink{webhose-free-news}{Free news} & 15~722 & 1~612 & 580 ± 600 & 12 ± 5 & news & topics \\ 
    \citelink{chen-etal-2021-dialogsum}{Dialogsum} & 12~812 & 0 & 122 ± \texttt{ }68 & ~ & casual & implicit \\ 
    \citelink{pang-lee-2005-seeing}{Rotten tomatoes} & 10~647 & 0 & 21 ± \texttt{ \ }9 & ~ & casual & implicit \\ 
    \citelink{szemraj-2023-goodreads}{Goodreads book genres} & 7~165 & 0 & 156 ± \texttt{ }89 & ~ & other & topics \\ 
    \citelink{burnham-etal-2024-political}{Political or not} & 2~174 & 1~681 & 245 ± \texttt{ }98 & ~ & other & topics \\ 
    \bottomrule
    \end{tabular}
    \caption{The politicalness datasets, ordered by size. The label source: implicit means the texts are typically and overwhelmingly non-political; topics have been labeled as described in section \ref{sec:methodology_data_preprocessing_politicalness}.}
    \label{fig:datasets_politicalness}
\end{table*}

\subsection{Dataset intersection}\label{sec:methodology_dataset_intersection}

When working with multiple datasets, it is crucial to check whether any of them intersect -- share some number of identical examples. It is important to determine the size of these intersections and adapt the research accordingly. If the overlap between two datasets is not negligible (below 10~\% of their size), it will affect training and evaluation of models. Using one of the datasets to evaluate a model trained on the other one will yield overoptimistic and misleading results, because the evaluation set will likely contain examples which the model has already encountered during training. We identify the overlapping datasets and exclude them from experiments where this spillover could occur.

The intersections of texts between every pair of datasets have been measured by comparing the titles and bodies. Beforehand, both the titles and the bodies have been stripped of any non-letter characters (even whitespace) and converted to lowercase, so that irrelevant text discrepancies would not manifest.

The preferable option is to simply compare the titles for an exact match. However, some rows or even whole datasets do not contain a title field, so it becomes necessary to compare the bodies in some way. A primitive char-to-char string comparison does not serve this purpose, because, for example, different web scrapers include different parts of the page, for instance the title of the comment section. Sophisticated string similarity algorithms (e.g. the Levenstein distance) are not a viable option, since they have high time complexities for such a large amount of text. A simpler algorithm has been chosen: slicing a middle part (50 characters or less if the body is shorter) out of the first dataset's row's body and testing, whether the second dataset's row's body contains it. Slicing the middle is considered optimal, as that is the part with the least probability of a scraping mismatch. Still, this method only gives an estimate of the true intersection -- for example, there are instances of false positive matches caused by two texts both citing one politician's statement in their middle.

In-depth details are captured in appendix~\ref{sec:appendix_methodology_dataset_intersection}.

\subsection{Model evaluation}\label{sec:methodology_model_evaluation}

All the validation and test set evaluations (of both existing and new custom models) on the available datasets have been performed using the following methodology.

The validation and test sets are created by sampling the whole datasets to either some fixed number or some fraction of examples. Using a fraction causes the size of each subset to be different, depending on the source dataset size. That would be an issue if they were to be combined afterwards into a single validation or test set, because the dataset portions contained in it would be highly unbalanced. Therefore, we never combine such subsets and instead we evaluate the models on each of them separately and, if it is desired and appropriate, we calculate the average of the individual metric results. This way, each subset affects the average equally, regardless of its (or the original dataset's) size.

The sampling always ensures an even distribution of the classes present in the sample and it is systematic (taking rows at regular intervals) by the body length in each class. If the number of rows in a dataset is less than the sample size, then as many rows as possible are taken from each class while still maintaining an even distribution of the classes. Then, the accuracy, precision, recall, $F_1$ score and confusion matrix of the model are measured. When a political leaning model does not support the center leaning class, it is only evaluated on left and right leaning texts.

\subsection{Existing model evaluation}

The models have been evaluated using the methodology prescribed above (section \ref{sec:methodology_model_evaluation}), while the datasets have been sampled to 1~000 rows each.

\subsubsection{Political leaning}

DistilBERT-PoliticalBias and Political ideologies RoBERTa fine-tuned are in fact trained to predict liberal vs. conservative leaning. That output has been mapped to left vs. right respectively. The same applies for DistilBERT political tweets, which is trained to predict democratic vs. republican leaning. 5 of the models do support the center leaning class, the other 4 only distinguish the left and right. DistilBERT-PoliticalBias uses a 5-level spectrum, which has been reduced to a 3-level one by mapping both highly and mildly left-leaning to the single left-leaning class and the same for right-leaning. Political DEBATE is an NLI classifier and has been supplied a simple zero-shot hypothesis \begin{spverbatim}This text supports {left | center | right} political leaning.\end{spverbatim}

According to their Hugging Face model cards, some of the models have been trained on some of the collected datasets. This information is captured in table \ref{fig:models_existing_leaning}.

\begin{table*}[!ht]
    \centering
    \small
    \begin{tabular}{lll}
        \toprule
        name & trained on & supports center class \\
        \midrule
        \citelink{bucket-2023-political}{Political bias BERT} & Article bias prediction & \cmark \\ 
        \citelink{sahitaj-2024-political}{Political bias prediction AllSides DeBERTa} & Article bias prediction & \cmark \\ 
        \citelink{jones-2024-political-bias}{DistilBERT-PoliticalBias} & GPT-4 political bias & \cmark \\ 
        \citelink{shrimali-2024-distilbert}{DistilBERT political fine-tune} & N/A & \cmark \\ 
        \citelink{burnham-etal-2024-political}{Political DEBATE large v1.0} & Dem., rep. party platform topics; Political tweets & \cmark \\ 
        \citelink{velez-2024-bert}{BERT political bias fine-tune} & N/A & \xmark \\ 
        \citelink{nayak-2024-roberta}{Political ideologies RoBERTa fine-tuned} & GPT-4 political ideologies & \xmark \\ 
        \citelink{palmqvist-2024-deberta}{DeBERTa political classification} & N/A & \xmark \\ 
        \citelink{newhauser-2022-distilbert}{DistilBERT political tweets} & a subset of Political tweets & \xmark \\ 
        \bottomrule
    \end{tabular}
    \caption{The existing political leaning models.}
    \label{fig:models_existing_leaning}
\end{table*}

\subsubsection{Politicalness}\label{sec:methodology_existing_evaluation_politicalness}

We have evaluated each of the models on each dataset separately. However, these results might end up to be misleading, because most of the datasets contain either only political (the political leaning datasets) or only non-political examples -- there are only a few featuring both classes (shown by table \ref{fig:datasets_politicalness}). In such circumstances, where the classes in each dataset are drastically unbalanced (or some are missing entirely), the metrics (accuracy, precision, recall and $F_1$ score) give deceptive results and can even be undefined due to division by zero. Therefore, we also provide an evaluation on an aggregate dataset, composed of samples (1~000 examples) from all the available datasets, which has a balanced class distribution (49~\% / 51~\%).

Political DEBATE is an NLI classifier and has been supplied a simple zero-shot hypothesis \begin{spverbatim}This text {is not | is} about politics.\end{spverbatim}

\subsection{Dataset benchmarks}\label{sec:methodology_dataset_benchmarks}

We have fine-tuned several sets of models to compare the character of the datasets and their suitability for training. The benchmark training sets have been relatively small (could have been even 5 times larger). Also (if not stated otherwise), no rigorous search for optimal hyperparameters has been done and the defaults provided by the Hugging Face transformers library \cite{wolf-2020-hugging-face} have been used with minimal adjustments. This is because the goal of these benchmarks was not to develop a well performing model, but rather to acquire information about the datasets themselves and to prove our hypotheses regarding general model performance.

We have diverted from the Hugging Face defaults by decreasing the maximum input length from 512 to 256 tokens. For reference, one token from a model tokenizer is on average equal to roughly one word. Any tokens past this limit get cut off -- the text is truncated from the right side. This decrease allows for larger batch sizes, cuts the training time in half and does not measurably affect the performance. Our explanation is that in most cases the texts clearly express their political views, basically from the beginning -- only opinions baked in some sort of a later punchline, past the maximum length, could hide from the model's sight. Refer to tables~\ref{fig:datasets_leaning} and~\ref{fig:datasets_politicalness} for the average number of words in the datasets.

Additionally, it was necessary to enable warmup steps. Otherwise, the models would sometimes learn to only predict a single class and never any other. We have not dived very deep into this issue, but since the learning rate is highest at the beginning of the training when not using warmup, we supsect that the optimizer can drastically overshoot the weights into a non-recoverable state. Gradually increasing the learning rate from zero to the maximum (the purpose of warming up) prevents this from happening. We have settled for a warmup ratio of 15~\% of the total steps.

We should also note that the batch size is selected automatically and in the case of our hardware (details in appendix \ref{sec:appendix_hardware}), it has been set to 8.

Several pretrained transformer language models have been tested and compared, namely BERT base (cased), RoBERTa base, DeBERTa~V3 base and POLITICS.

\subsubsection{Leave-one-in}\label{sec:methodology_benchmark_leave_one_in}

From each dataset, a sample of 2~000 rows (or less if the dataset is smaller) has been taken while ensuring an equal distribution of classes and systematically sampling by the body length in each class. A model has been fine-tuned separately on each of these samples. Before that, 15~\% of the dataset has been taken as a test set and another 15~\% as a validation set while following the methodology prescribed above (in section \ref{sec:methodology_model_evaluation}).

This approach is essentially the way most of the existing models have been produced -- training and evaluating on a single dataset and thus only on a single type of text.

This benchmark has not been conducted on datasets for politicalness, as most of them contain purely one class, and so training a model on them is impossible.

\subsubsection{Leave-one-out}\label{sec:methodology_benchmark_leave_one_out}

If two datasets intersect with each other (as measured with the methodology described in section \ref{sec:methodology_dataset_intersection}) by a large amount, then the smaller one is not a part of this benchmark to avoid spillover. This applies for two political leaning datasets: Webis bias flipper 18 and Webis news bias 20, which intersect significantly with each other and with Article bias prediction (section \ref{sec:results_dataset_intersection}).

From each dataset, a sample (2~000 rows for political leaning, 1~000 for politicalness -- or less if the dataset is smaller) has been taken while ensuring an equal distribution of classes and systematically sampling by the body length in each class. Several models have been fine-tuned separately, each on a concatenation of all but one (the left-out dataset) of these samples. This has been repeated, leaving out a different dataset at a time. The most interesting evaluation is then on this left-one-out dataset, because it shows how well the model handles not just unseen individual texts, but also different text styles or topics from a different time period. This is why the validation set in this benchmark consists purely of examples from the left-out dataset, so that the best model checkpoint can be selected based on that performance. The test set size has been chosen to 15~\% of each dataset. A fraction instead of a fixed size has the advantage of not depleting the small datasets, but also making use of the greater amount of data in the large ones, where we can afford it. The validation set size has been set to 1~000 (or less if the dataset is smaller) of the remaining examples. When creating both the validation and the test sets, we have followed the methodology prescribed above (in section \ref{sec:methodology_model_evaluation}).

The fact that some political leaning datasets do not contain any center leaning examples has consequences on the class distribution of the concatenation: taking an equal number of rows from each dataset results in the center class being underrepresented. Our experiments have revealed that the training class distribution is crucial for the model not to over-prioritize the left and right classes. Utilization of class weights does not resolve this issue. And, in fact, the classes need to be balanced almost equally. For this we have added a multiplier which increases the number of taken center leaning examples from the dataset that have them. We adjust it to even out the class distribution.

We have also conducted this benchmark once more with a second goal: to find the optimal training setup that would produce a model for classifying political leaning (for politicalness, we consider the performance of the best existing model sufficient as a filter of non-political texts) that would be able to handle out-of-distribution texts. Our assumption is that a new model trained later by following this exact setup on all the available datasets combined is going to be as versatile as possible, since the methodology will be proven by the strict and tough leave-one-out test. Such a model should not only perform well on all the data we have collected, but also on unseen data or even texts yet to be written in the future. We have taken a larger training sample, selected the most effective pretrained model, and focused on optimizing its performance by tuning hyperparameters.

10~000 has turned out to be a good sample size to be taken from each dataset for training, since it still makes it possible to balance out the classes (with the center multiplier). The total number of examples in the training concatenation came out to be around 100~000.

POLITICS has proven to be the best model for further fine-tuning, consistently scoring the highest in all previous benchmarks. We have inspected the configuration proposed in the paper released alongside the model \cite{liu-etal-2022-politics} and explored works discussing generally the most influential hyperparameters when fine-tuning BERT and its derivatives for text classification \cite{chi-sun-etal-2019-how-to, lorenzoni-etal-2024-exploring}. Based on these recommendations, we have run the Optuna hyperparameter search \cite{akiba-etal-2019-optuna} in the following search space ([] denotes a closed interval, \{\} a predefined set and the values following $\rightarrow$ are the final ones selected as most optimal):

\begin{enumerate}
    \item Attention dropout: [0.1, 0.3] $\rightarrow$ 0.1106.
    \item Hidden layers dropout: [0.1, 0.5] $\rightarrow$ 0.2212.
    \item Classifier layer dropout: [0.0, 0.5] $\rightarrow$ 0.07925.
    \item Learning rate: [1e-5, 7e-5], sampled logarithmically $\rightarrow$ 5.97e-5.
    \item Warmup ratio: [0.05, 0.25] $\rightarrow$ 0.20885.
    \item Batch size: \{8, 16, 32, 48, 64\} $\rightarrow$ 64.
    \item Weight decay: [0.0001, 0.1], sampled logarithmically $\rightarrow$ 0.00015.
\end{enumerate}

The number of epochs for each trial has been set to 5 and the best performing checkpoint from the training was always selected for comparing the trials.

For a training set of such size, it may also become beneficial to employ a larger model. POLITICS, however, only comes in a base variant, and so we have decided to experiment with fine-tuning DeBERTa~V3 large, since its architecture is the cutting-edge among currently available transformer language models. A hyperparameter search was no longer an option, as this model is much more memory demanding and takes a long time to fine-tune on the hardware available for this research (details in appendix \ref{sec:appendix_hardware}). The hyperparameters were therefore needed to be tweaked manually. We have arrived at these final adjustments to the Hugging Face defaults:

\begin{enumerate}
    \item Maximum input length: 512 $\rightarrow$ 256 tokens. The reasons described in section \ref{sec:methodology_dataset_benchmarks}.
    \item Learning rate: 5e-5 $\rightarrow$ 3e-5. Increasing it can result in the model overshooting loss minima and makes the training unstable. Decreasing causes slow convergence and potentially stalling in a local loss minimum.
    \item Warmup ratio: 0 $\rightarrow 0.1$. The reasons described in section \ref{sec:methodology_dataset_benchmarks}.
    \item Batch size: 8 $\rightarrow$ 40. The GPU's VRAM has only allowed for the true batch size of 10, so it had to be simulated by setting gradient accumulation steps to 4. Small batches resulted in much longer training times and larger ones made the learning curve more stable, but values too high harmed the peak performance the model was able to achieve.
    \item Number of epochs: 3 $\rightarrow$ 4. While the model performance has usually peaked before the end of the third epoch, in some cases it was necessary to train for a little longer.
\end{enumerate}

Tuning other hyperparameters has not brought significant improvements.

\begin{figure*}[ht]
    \centering
    \includegraphics[width=\textwidth]{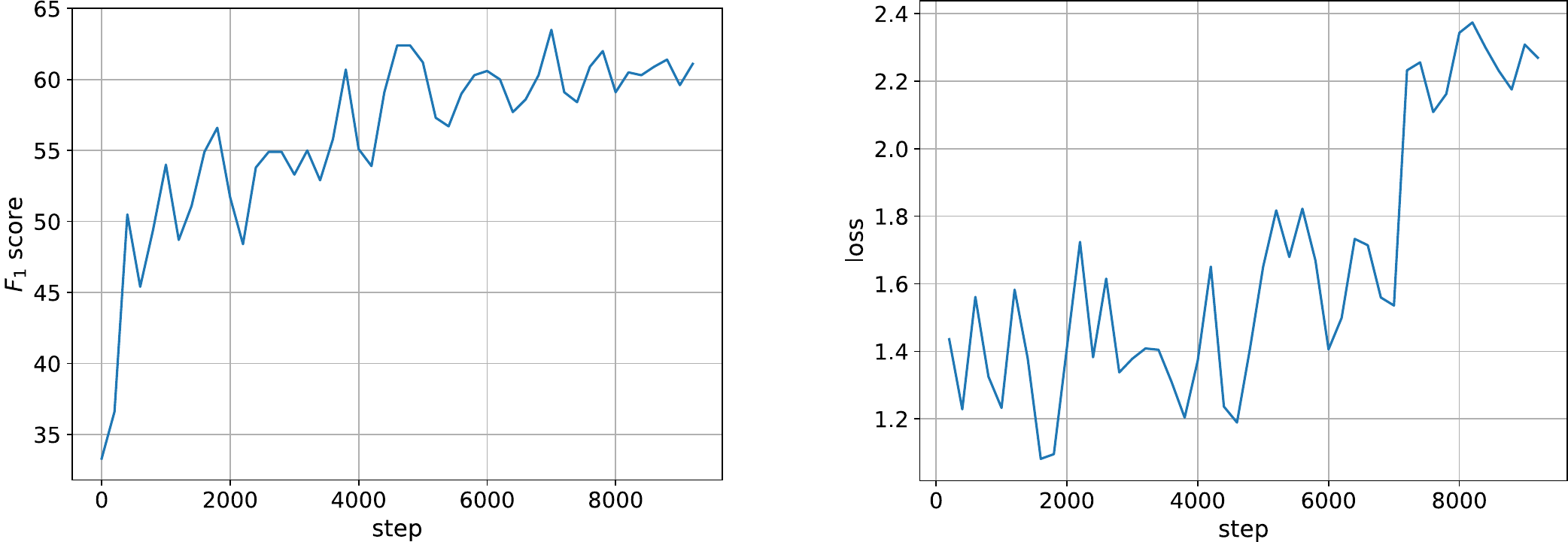}
    \caption{The progression of the $F_1$ score and loss on the validation set across training steps when fine-tuning DeBERTa~V3 large for the leave-one-out benchmark with 10~000 examples from each dataset, leaving out and validating on the Article bias prediction dataset. The $F_1$ score peaks at around 7~000 steps -- at the same point as the loss drops for the last time before starting to dramatically increase, indicating overfitting.}
    \label{fig:leave_one_out_benchmark_eval_loss_f1}
\end{figure*}

Both in the hyperparameter search for POLITICS and while tweaking DeBERTa large, we have chosen the Article bias prediction to be the left-out dataset and aimed for the best performance when evaluating the model on it. There are multiple reasons for this choice: it features the center leaning class, has texts annotated individually (not by the overall leaning of the media outlet) and covers a wide variety of topics. The logging interval has been set to 200 steps and $F_1$ score has been the evaluation metric based on which we have selected the best model checkpoints. Among these, we have also inspected and considered the confusion matrices. Figure \ref{fig:leave_one_out_benchmark_eval_loss_f1} shows an example of the training progress.

Using the selected optimized hyperparameters with POLITICS and DeBERTa large, we have conducted the whole leave-one-out benchmark again, leaving a different dataset at a time.

\subsection{Custom model training}

After optimizing the training setup for out-of-distribution performance in the leave-one-out dataset (section \ref{sec:methodology_benchmark_leave_one_out}), we have proceeded to train a model on all the available datasets. We have employed the exact same setup, only this time not leaving any dataset out -- except for Webis bias flipper 18 and Webis news bias 20 (not included in the leave-one-out benchmark either for the reasons described there). It was also necessary to change the validation and test set creation -- from each dataset, 15~\% of examples has been taken as the test set. As the validation set, a concatenation of 100 examples taken from each of the datasets has been used -- in this case, choosing a fixed size of the validation set samples is better, because they need to be balanced. Otherwise, the selection of the best model checkpoint would be influenced more by the larger datasets.

\section{Results}

\subsection{Dataset intersection}\label{sec:results_dataset_intersection}

No notable intersections have been detected among the politicalness datasets.

Of the political leaning datasets, only Webis bias flipper 18 and Webis news bias 20 intersect significantly with each other -- by 5~132 instances (equal to 80.2~\% and 66.5~\% of the total examples in each dataset, respectively). They both also overlap notably with Article bias prediction -- 2~716 and 3~824 matches have been found, respectively. This means that models (both existing and our new) trained on one of these are going to have boosted performance on the other two as well -- not only because they contain the same topics, but especially if some articles contained in a training set of one dataset would also make their way into a test set from one of the other two datasets. This spillover is something we have avoided in the leave-one-out benchmark (\ref{sec:methodology_benchmark_leave_one_out}).

The complete resulting tables of the intersections are recorded in appendix \ref{sec:appendix_dataset_intersection}.

\subsection{Existing models evaluation}

\subsubsection{Political leaning}

The resulting $F_1$ scores are recorded in table \ref{fig:models_eval_existing_leaning}. The best performing classifier is Political bias prediction AllSides DeBERTa, scoring the highest on both in- and out-of-distribution data. This model is therefore what we consider the state-of-the-art among the currently available models. DistilBERT political tweets does have slightly higher average scores, however since it does not support the center leaning class, it has a statistical advantage of choosing 1 out of 2 classes (50~\%) instead of 3 (33~\%).

\begin{table*}[ht]
    \centering
    \small
    \begin{tabular}{lrrrrr|rrrr}
    \toprule
    ~ & \rotatebox[origin=l]{90}{Political bias BERT} & \rotatebox[origin=l]{90}{Political bias prediction AllSides DeBERTa} & \rotatebox[origin=l]{90}{DistilBERT-PoliticalBias} & \rotatebox[origin=l]{90}{DistilBERT political fine-tune} & \rotatebox[origin=l]{90}{Political DEBATE large v1.0} & \rotatebox[origin=l]{90}{BERT political bias fine-tune} & \rotatebox[origin=l]{90}{Political ideologies RoBERTa fine-tuned} & \rotatebox[origin=l]{90}{DeBERTa political classification} & \rotatebox[origin=l]{90}{DistilBERT political tweets} \\ 
    supports center leaning & \cmark & \cmark & \cmark & \cmark & \cmark & \xmark & \xmark & \xmark & \xmark \\
    \midrule
    Article bias prediction & \textcolor{orange}{57} & \textcolor{orange}{65} & 17 & 26 & 51 & 35 & 43 & 61 & 57 \\ 
    BIGNEWSBLN & 37 & 54 & 18 & 24 & 42 & 40 & 46 & 62 & 57 \\ 
    CommonCrawl news articles & 46 & 68 & 18 & 28 & 54 & 40 & 44 & 65 & 56 \\ 
    Dem., rep. party platform topics & 50 & 60 & 42 & 42 & \textcolor{orange}{62} & 38 & 58 & 57 & 62 \\ 
    GPT-4 political bias & 22 & 15 & \textcolor{orange}{82} & 39 & 57 & 26 & 81 & 46 & 73 \\ 
    GPT-4 political ideologies & 38 & 75 & 54 & 63 & 82 & 34 & \textcolor{orange}{99} & 63 & 81 \\ 
    Media political stance & 43 & 73 & 34 & 43 & 56 & 41 & 51 & 68 & 61 \\ 
    Political podcasts & 26 & 80 & 34 & 46 & 69 & 50 & 53 & 74 & 71 \\ 
    Political tweets & 41 & 57 & 42 & 55 & \textcolor{orange}{57} & 47 & 58 & 46 & \textcolor{orange}{74} \\ 
    Qbias & 29 & 46 & 23 & 21 & 37 & 47 & 47 & 55 & 53 \\ 
    Webis bias flipper 18 & \textcolor{orange}{38} & \textcolor{orange}{59} & 16 & 24 & 42 & 37 & 47 & 58 & 53 \\ 
    Webis news bias 20 & \textcolor{orange}{48} & \textcolor{orange}{75} & 16 & 21 & 39 & 39 & 47 & 59 & 56 \\ 
    \midrule
    \textcolor{orange}{in-distribution} average & 48 & 66 & 82 & -- & 60 & -- & 99 & -- & 74 \\ 
    out-of-distribution average & 37 & \textbf{59} & 29 & -- & 53 & -- & 52 & -- & 62 \\ 
    overall average & 40 & \textbf{61} & 33 & 36 & 54 & 40 & 56 & 60 & 63 \\ 
    \bottomrule
    \end{tabular}
    \caption{Average $F_1$ scores of evaluating the existing political leaning models on all available datasets, sampled to 1~000 examples each. To avoid misinterpreting them, models with different numbers of supported classes should be compared separately -- a totally randomly predicting model choosing between two classes would statistically achieve an accuracy of around 50~\%, but in the case of three classes that baseline would be 33~\%. We observe that all models perform consistently and significantly better on \textcolor{orange}{in-distribution} datasets -- the ones they have been trained on (recorded also in table \ref{fig:models_existing_leaning}).}
    \label{fig:models_eval_existing_leaning}
\end{table*}

\subsubsection{Politicalness}\label{sec:results_existing_evaluation_politicalness}

The most important results of the evaluation -- on the aggregate dataset (described in section \ref{sec:methodology_existing_evaluation_politicalness}) -- are recorded in table \ref{fig:models_eval_existing_politicalness}. The complete resulting table of each model evaluated on each dataset separately can be found in appendix \ref{sec:appendix_existing_evaluation_politicalness}.

\begin{figure}[tbh]
    \centering
    \includegraphics[width=\columnwidth]{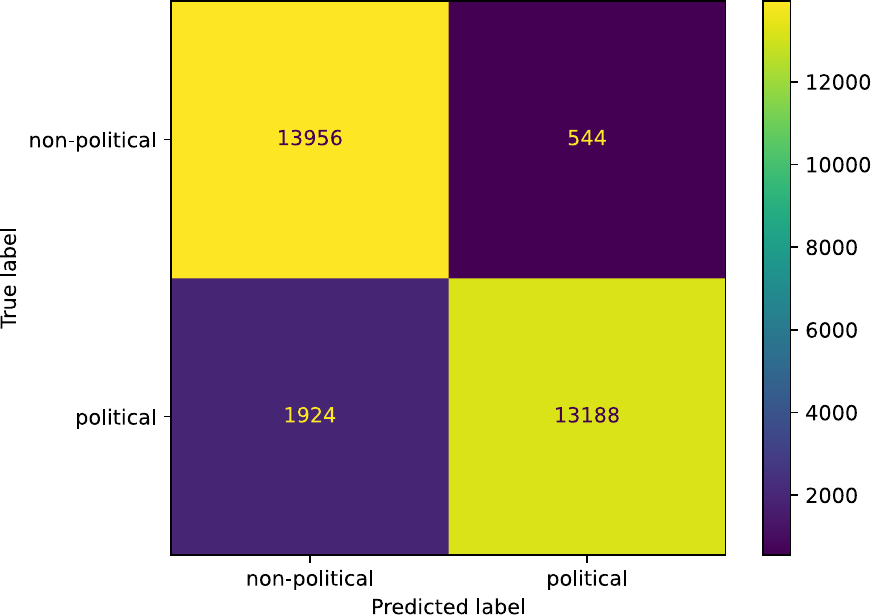}
    \caption{The resulting confusion matrix of Political DEBATE large classifying by politicalness evaluated on an aggregate dataset consisting of samples (1~000 examples each) from all available (both politicalness and political leaning) datasets.}
    \label{fig:politicalness_debate_confusion_aggregate}
\end{figure}

Political DEBATE performs the best and is accurate enough for real world application tasks. Its confusion matrix (figure \ref{fig:politicalness_debate_confusion_aggregate}) displays a greater number of false negatives than false positives, which indicates the model is more prone to classifying texts that actually are about politics as non-political than the other way around. This is to be considered when deploying to a real application, e.g. as a sieve in front of a political leaning classifier -- whether it is more desirable to strictly filter out non-political texts, or to aim for processing as many texts as possible, allowing some amount of them to be potentially non-political and thus irrelevant. This tuning could be implemented by adjusting the threshold for the model's output confidence score -- as done by us in section \ref{sec:methodology_data_preprocessing_leaning}. However, this particular model may not be well-suited for this approach in an application outside of scientific research, because the distribution of its confidence scores is extremely unbalanced -- in 90~\% of cases, the score exceeds 0.99.

\begin{table}[ht]
    \centering
    \small
    \begin{tabular}{lrr}
    \toprule
    model & accuracy & $F_1$ score \\
    \midrule
    Classifier main subject politics & 69.5 & 66.9 \\
    Political DEBATE large & \textbf{90.4} & \textbf{90.4} \\
    Topic politics & 76.3 & 75.1 \\
    \bottomrule
    \end{tabular}
    \caption{Results of evaluating the existing politicalness models on an aggregate dataset consisting of samples (1~000 examples each) from all available (both politicalness and political leaning) datasets.}
    \label{fig:models_eval_existing_politicalness}
\end{table}

\begin{table*}[!ht]
    \centering
    \small
    \begin{tabular}{lrrrrrrrr}
    \toprule
    \textsc{the left-in dataset} $\downarrow$ & \multicolumn{2}{c}{BERT} & \multicolumn{2}{c}{RoBERTa} & \multicolumn{2}{c}{DeBERTa} & \multicolumn{2}{c}{POLITICS} \\ 
    \textsc{test set distribution} $\rightarrow$ & left-in & unseen & left-in & unseen & left-in & unseen & left-in & unseen \\ 
    \midrule
    \multicolumn{5}{l}{\textsc{with the center leaning class}} \\
    Article bias prediction & 64.8 & 39.6 & 58.9 & 33.4 & 70.8 & 37.4 & 81.3 & 51 \\ 
    BIGNEWSBLN & 73.3 & 35.9 & 77.9 & 32.6 & 80.9 & 40.6 & 89.7 & 38.4 \\ 
    CommonCrawl news articles & 74.4 & 33.2 & 78.7 & 37.2 & 78.9 & 35.5 & 82.2 & 49 \\ 
    GPT-4 political bias & 90.1 & 35.4 & 91.2 & 37.8 & 92.4 & 36.6 & 91.3 & 37 \\ 
    Qbias & 44.3 & 38.4 & 53.6 & 41.5 & 38.4 & 34.7 & 51.1 & 48.3 \\ 
    Webis bias flipper 18 & 77.5 & 41.6 & 82.1 & 33.6 & 82.3 & 45.4 & 89 & 50.8 \\ 
    Webis news bias 20 & 74.4 & 40.9 & 80 & 36.9 & 70.6 & 42.1 & 83.3 & 50.3 \\ 
    \midrule
    average & 71.3 & 37.9 & 74.6 & 36.1 & 73.5 & 38.9 & 81.1 & 46.4 \\ 
    \midrule
    \multicolumn{5}{l}{\textsc{without the center leaning class}} \\
    Dem., rep. party platform topics & 76.6 & 45 & 75.4 & 53.3 & 57.8 & 38.7 & 76.8 & 60 \\ 
    GPT-4 political ideologies & 99 & 53.8 & 99.2 & 53.5 & 99.2 & 56.5 & 98.7 & 57.1 \\ 
    Media political stance & 83.7 & 49.2 & 87.9 & 50.7 & 82.1 & 55.3 & 90.3 & 67.5 \\ 
    Political podcasts & 99.5 & 43.4 & 99.6 & 44.5 & 99.8 & 45 & 99.5 & 62.5 \\ 
    Political tweets & 71.6 & 54.6 & 73.7 & 57.7 & 65.8 & 51.4 & 74.8 & 66.7 \\ 
    \midrule
    average & 86.1 & 49.2 & 87.2 & 51.9 & 80.9 & 49.4 & 88 & 62.8 \\ 
    \bottomrule
    \end{tabular}
    \caption{Resulting $F_1$ scores on political leaning classification after evaluating the base versions of the BERT (cased), RoBERTa, DeBERTa~V3 and POLITICS models (test set size: 15~\%; train set size: 2~000 from each dataset) fine-tuned on one (the left-in) of the available datasets (10) at a time (the rows) -- as described in section \ref{sec:methodology_benchmark_leave_one_in}. The values in the left-in distribution columns are the score on the single dataset the model has been trained on. The unseen columns are the average score on all the other datasets. To avoid misinterpreting the results, models with different numbers of supported classes should be compared separately -- a totally randomly predicting model choosing between two classes would statistically achieve an accuracy of around 50~\%, but in the case of three classes that baseline would be 33~\%. We observe that all the classifiers perform consistently and significantly better on the datasets they have been trained on (in-distribution).}
    \label{fig:models_eval_in_leaning}
\end{table*}

\subsection{Leave-one-in dataset benchmark}

Table \ref{fig:models_eval_in_leaning} displays the resulting $F_1$ scores. As with the existing models -- trained in a similar manner (on just one dataset) -- these benchmarking prototype models also break on unseen data.

\subsection{Leave-one-out dataset benchmark}

Again, these results confirm the phenomenon of the models' performance dropping on unfamiliar data. The out-of-distribution scores are generally higher the newer the architecture of the model. However, POLITICS which is based on RoBERTa is clearly the best.

\subsubsection{Political leaning}\label{sec:results_dataset_benchmark_out_leaning}

The $F_1$ scores are shown in table \ref{fig:models_eval_out_leaning}. When evaluated on a left-out dataset that features all three leaning classes (does not miss the center), the models perform generally worse -- that is expected both statistically and because classifying the center-leaning texts might be harder.

Compared to POLITICS trained on just 2~000 examples from each dataset (table \ref{fig:models_eval_existing_leaning}), DeBERTa large with manually tweaked hyperparameters actually performs 7.8~\% lower on in-distribution data, but scores the best of all models on out-of-distribution, so this training has achieved its desired goal.

\begin{table*}[tbp]
    \centering
    \small
    \begin{tabular}{lrrrrrrrr}
    \toprule
    \textsc{the left-out dataset} $\downarrow$ & \multicolumn{2}{c}{BERT} & \multicolumn{2}{c}{RoBERTa} & \multicolumn{2}{c}{DeBERTa} & \multicolumn{2}{c}{POLITICS} \\ 
    \textsc{test set distribution} $\rightarrow$ & trained & left-out & trained & left-out & trained & left-out & trained & left-out \\ 
    \midrule
    Article bias prediction & 77.7 & 47 & 80.1 & 48.4 & 78.6 & 49.3 & 83.4 & 61.4 \\ 
    BIGNEWSBLN & 77.5 & 39.3 & 80.7 & 38.1 & 77.1 & 42 & 83.1 & 56.8 \\ 
    CommonCrawl news articles & 77 & 38.7 & 80.1 & 51 & 77.4 & 50 & 83.7 & 51.6 \\ 
    Dem., rep. party platform topics & 76.6 & 59.6 & 80.3 & 62.4 & 78.2 & 63.1 & 83.8 & 66.4 \\ 
    GPT-4 political bias & 74.4 & 42 & 78.7 & 45.8 & 76.3 & 37.2 & 82.7 & 44 \\ 
    GPT-4 political ideologies & 74 & 73.5 & 77.9 & 83.2 & 73 & 79 & 81.1 & 86.3 \\ 
    Media political stance & 73.1 & 54.4 & 78.3 & 57.3 & 76.7 & 64.9 & 82.3 & 66.5 \\ 
    Political podcasts & 73.4 & 61.8 & 77 & 55.9 & 48.5 & 55.1 & 81.2 & 76.6 \\ 
    Political tweets & 76.6 & 41 & 80.6 & 40.9 & 76.7 & 43.7 & 84.4 & 55.4 \\ 
    Qbias & 78.9 & 36.9 & 83.4 & 38.3 & 80.5 & 41.6 & 86.2 & 42.6 \\
    \midrule
    average & 75.9 & 49.4 & 79.7 & 52.1 & 74.3 & 52.6 & 83.2 & 60.8 \\ 
    \bottomrule
    \end{tabular}
    \vfill
    \vspace{0.5cm}
    \begin{tabular}{lrrrrrrrr}
    \toprule
    \textsc{the left-out dataset} $\downarrow$ & \multicolumn{2}{c}{POLITICS} & \multicolumn{2}{c}{DeBERTa large} \\ 
    \textsc{test set distribution} $\rightarrow$ & trained & left-out & trained & left-out \\ 
    \midrule
    Article bias prediction & 84.5 & 63.1 & 78 & 64.3 \\ 
    BIGNEWSBLN & 84.4 & 60.6 & 76.4 & 52 \\ 
    CommonCrawl news articles & 84.2 & 58 & 78.6 & 45.9 \\ 
    Dem., rep. party platform topics & 85.9 & 70 & 72.9 & 75 \\ 
    GPT-4 political bias & 84.1 & 45.9 & 76 & 46 \\ 
    GPT-4 political ideologies & 84 & 86.5 & 62.3 & 96.2 \\ 
    Media political stance & 84.3 & 72.7 & 78.2 & 77.9 \\ 
    Political podcasts & 83.1 & 79.1 & 74.6 & 91.2 \\ 
    Political tweets & 86.1 & 62 & 75.8 & 71.3 \\ 
    Qbias & 88.6 & 42.9 & 80.9 & 42.3 \\ 
    \midrule
    average & 84.9 & 64.1 & 75.4 & 66.2 \\ 
    \bottomrule
    \end{tabular}
    \caption{Resulting $F_1$ scores on political leaning classification after evaluating the base versions of the BERT (cased), RoBERTa, DeBERTa~V3 and POLITICS models (test set size: 15~\%; train set size: 2~000 from each dataset) and then POLITICS and DeBERTa~V3 large with optimized hyperparameters (test set size: 15~\%; train set size: 10~000 from each dataset) fine-tuned on all available datasets (10) except for one (the left-out) when leaving out a different dataset at a time (the rows) -- as described in section \ref{sec:methodology_benchmark_leave_one_out}. The values in the trained distribution columns are the average score on all the datasets the model has been trained on. The left-out columns are the score on the one left-out dataset. We observe that all the classifiers perform consistently and significantly better on the datasets they have been trained on (in-distribution).}
    \label{fig:models_eval_out_leaning}
\end{table*}

\begin{figure*}[bht]
    \centering
    \includegraphics[width=\textwidth]{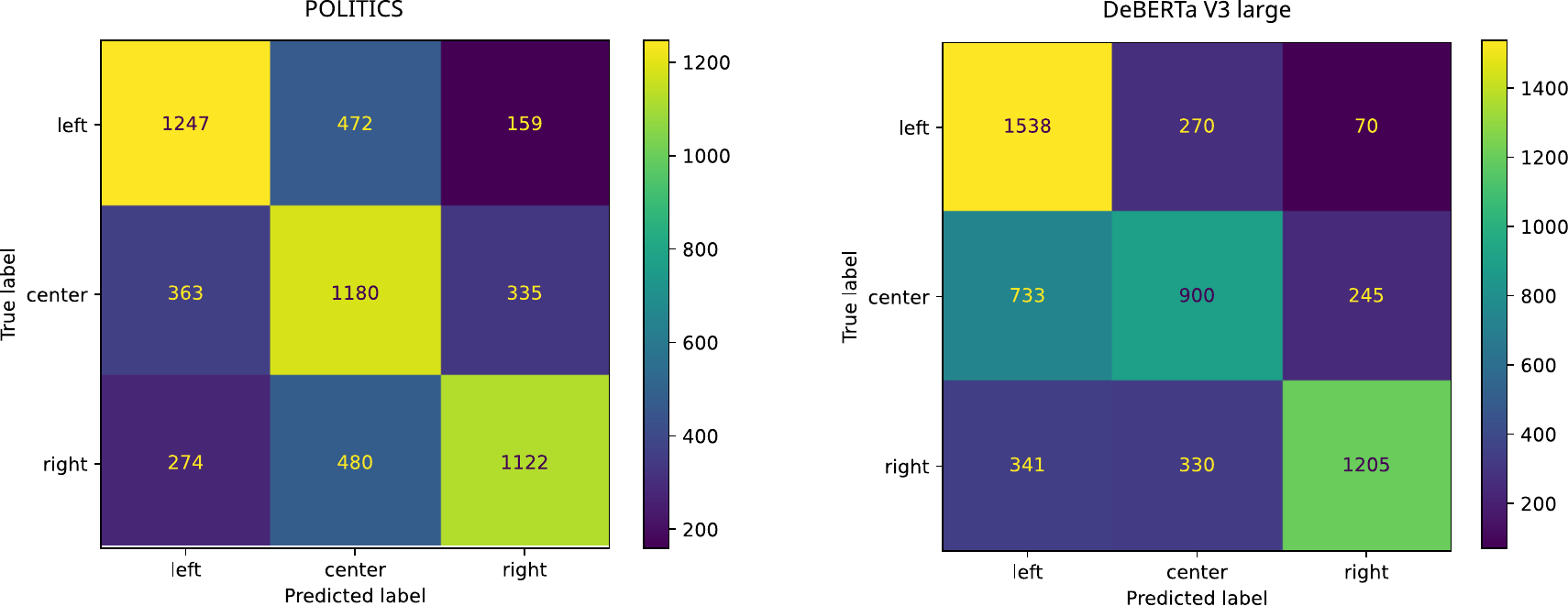}
    \caption{Confusion matrices of POLITICS and DeBERTa~V3 large with optimized hyperparameters (test set size: 15~\%; train set size: 10~000 from each dataset) fine-tuned on all available datasets (10) except for one (the left-out) -- in this case, Article bias prediction -- as described in section \ref{sec:methodology_benchmark_leave_one_out}. These matrices show the models' predictions on the test set consisting of just the left-out dataset.}
    \label{fig:leave_one_out_benchmark_confusion}
\end{figure*}

We also include example confusion matrices of the optimized models in figure \ref{fig:leave_one_out_benchmark_confusion}.

\subsubsection{Politicalness}\label{sec:results_dataset_benchmark_out_politicalness}

Table \ref{fig:models_eval_out_politicalness} captures the $F_1$ scores. It does not contain the results on individual datasets (28), as that would make it very large. We include those in appendix \ref{sec:appendix_leave_one_out_politicalness}. Nevertheless, this simplified table is enough to show that the models again score higher on familiar datasets. While consistent, this effect is not as drastic as with the task of political leaning. What also makes it less troubling is that the scores are overall very high, even on unseen types of texts, which as well indicates that this task is significantly easier than classification of political leaning.

\begin{table}[ht]
    \centering
    \small
    \begin{tabular}{lrr}
    \toprule
    \textsc{\textsc{test set distribution}} $\rightarrow$ & trained & left-out \\ 
    \midrule
    BERT & 97.6 & 93.3 \\ 
    RoBERTa & 95.9 & 91 \\ 
    DeBERTa & 97.8 & 92.9 \\ 
    POLITICS & 97.8 & 92.5 \\ 
    \bottomrule
    \end{tabular}
    \caption{Resulting $F_1$ scores on politicalness classification after evaluating the base versions of the BERT (cased), RoBERTa, DeBERTa~V3 and POLITICS models (test set size: 15~\%; train set size: 1~000 from each dataset) fine-tuned on all available datasets (28) except for one (the left-out) -- as described in section \ref{sec:methodology_benchmark_leave_one_out}. The values in the trained distribution column are the average score on all the datasets the model has been trained on. The left-out column is the score on the one left-out dataset. We observe that all the classifiers perform consistently better on the datasets they have been trained on (in-distribution).}
    \label{fig:models_eval_out_politicalness}
\end{table}

\subsection{Custom model training}

Our models trained on all available datasets using the methodology optimized for classifying unfamiliar data (described in section \ref{sec:methodology_benchmark_leave_one_out}) have achieved the $F_1$ scores recorded in table \ref{fig:custom_training_leaning}. They establish the new state-of-the-art performance across all the evaluated datasets.

\begin{table*}[!ht]
    \centering
    \small
    \begin{tabular}{lrrrr}
    \toprule
    ~ & \multicolumn{2}{c}{POLITICS} & \multicolumn{2}{c}{DeBERTa V3 large} \\
    ~ & accuracy & $F_1$ score & accuracy & $F_1$ score \\ 
    \midrule
    Article bias prediction & 84.7 & 84.6 & 89 & 89 \\ 
    BIGNEWSBLN & 89.2 & 89.2 & 88.6 & 88.6 \\ 
    CommonCrawl news articles & 85.2 & 85.1 & 88.9 & 88.9 \\ 
    Dem., rep. party platform topics & 77.6 & 77.3 & 85.5 & 85.6 \\ 
    GPT-4 political bias & 84.8 & 84.2 & 87 & 86.9 \\ 
    GPT-4 political ideologies & 97.5 & 97.5 & 99.6 & 99.6 \\ 
    Media political stance & 90.8 & 91.1 & 91.6 & 93.1 \\ 
    Political podcasts & 99.4 & 99.5 & 99.8 & 99.8 \\ 
    Political tweets & 76.4 & 76.2 & 82.1 & 82.1 \\ 
    Qbias & 51.6 & 51 & 58 & 57.9 \\ 
    \midrule
    average & 83.7 & 83.6 & 87 & 87.2 \\
    \bottomrule
    \end{tabular}
    \caption{Results of evaluating classifiers of political leaning trained on all available datasets (10) with optimized hyperparameters for out-of-distribution performance (as described in section \ref{sec:methodology_benchmark_leave_one_out}).}
    \label{fig:custom_training_leaning}
\end{table*}

\section{Discussion}\label{sec:discussion}

Our research hypothesis posits that models classifying political leaning perform well on the datasets they have been trained on (in-distribution), but exhibit significantly diminished precision when applied to texts of unseen styles or just written in a different time period (out-of-distribution). It is aligned with the existing literature: \cite{cohen-ruths-2021-classifying} demonstrated that the classifiers achieve great accuracy (>90~\%) on datasets of tweets written by politicians and politically vocal users, but their performance plummets to barely 65~\% when applied to regular users. Similarly, \cite{yan-etal-2017-perils} found that cross-domain generalizability was remarkably poor across datasets from congressional records, media websites, and wikis, despite strong within-dataset cross-validation results. Their work revealed that the difficulty stems not merely from technical limitations in transferability, but from fundamental differences in the nature of data across domains.

Our evaluation of the existing models does confirm these claims: Despite achieving impressive results on familiar datasets, their performance suffers on the unseen ones. Although this is normally somewhat expected, the drop in accuracy is not only consistent, but also significant (7~\% at minimum).

The methodology of our leave-one-in benchmark is very close to the way that existing works have approached this task: training and evaluating models on a single data dataset and thus a single type of texts. Our results have shown that it is relatively easy to achieve decent in-distribution performance, even with a small training sample. However, once exposed to unfamiliar datasets, these models fail drastically.

The leave-one-out benchmark has yielded very similar results, again supporting our hypothesis, despite training on a much larger and more diverse set. Inspecting the models trained on the increased sample size (10~000) and with adjusted hyperparameters, we notice a level of improvement: Though they are subject to a decrease in performance on the trained datasets, they achieve higher out-of-distribution scores. We consider this tradeoff a success -- it reduces the ubiquitous fall of accuracy on unseen data which all the existing and our previous prototype models suffer from. The results show that this configuration is suitable and appropriate for training on all available data, which should result in a model transferable to new unseen texts.

We have trained two such models (one based on POLITICS and second on DeBERTa~V3 large) for classification of political leaning. Their performance is the new state-of-the-art on all the collected datasets. However, the extent, to which our new models trained on all the currently available datasets are able to generalize across domains, can only be revealed by exposing them to datasets yet to be published or to handcrafted testing examples.

In general, when it comes to the prototype models trained for dataset benchmarks, the newer the pretrained language model architecture, the better the fine-tuned model performance: DeBERTa~V3 (2021) outperforms RoBERTa (2019) which outperforms BERT (2018). There is, however, an exception: POLITICS, based on RoBERTa, performs the best in all the conducted measurements. These results prove that continued training for specific text domains is effective and confirm that the effort of the authors \cite{liu-etal-2022-politics} has paid off.

We consider it important to also mention the performance of generative LLMs with hundreds of billions of parameters, as they have become the kitchen-sink approach to machine learning problems, and thus are a candidate for solving the tasks of this study as well. While they have demonstrated impressive capabilities in text classification tasks, their adoption in both scientific research and practical applications comes with significant drawbacks. These massive models require substantial computational resources, making them either expensive, or slow. Perhaps most concerning from an academic perspective is the predominantly closed-source nature of many state-of-the-art LLMs, which undermines the reproducibility of achieved results and the option to inspect the internal model behavior. As \cite{burnham-etal-2024-political} demonstrate with their Political DEBATE models (performing the best in our tests for the task of politicalness), domain-specialized smaller models can match or even outperform proprietary LLMs on text classification tasks while being orders of magnitude more efficient. For these reasons, we have decided to explore the fine-tuning of these, thousands of times smaller, language models.

\section{Limitations}

While this work attempts to cover its topics comprehensively, it has some limitations.

First, the classification of political leaning is simplified. Three classes may not model the real spectrum in sufficient detail. Increasing the number of classes is one way to enhance the resolution. Alternatively, using regression instead of classification could provide a more nuanced representation by placing texts on a continuous numerical scale that reflects varying degrees of political leaning. Furthermore, it is important to consider that some texts may simultaneously support both left- and right-leaning perspectives, suggesting the need for a ``mixed'' class or a method to predict the separate extents to which a text aligns with each side (similarly to the task of stance detection -- in favor / against).

Second, our definition of politicalness focuses solely on whether the main topic of a text is politics, ignoring whether it expresses a political opinion. For example, texts like Wikipedia articles or dictionary entries may discuss political topics without actually showing any bias. These texts should ideally get filtered away and not be allowed as inputs to a political leaning classifier, but with our simplified definition, they would. A more nuanced approach would require data with fine-grained labels which is not currently available.

Third, this study is limited by its focus on the U.S. political context, where the left-leaning class aligns predominantly with liberal and democratic views, and the right-leaning class with conservative and republican views. This framework does not translate well to broader politological contexts, or distinctions on multiple axes, such as conservative vs. progressive and authoritarian vs. liberal.

Fourth, our work explores neither data in other languages than English (which there is a lack of), nor multilingual models.

Fifth, the performance of the NLI classifier -- Political DEBATE -- is potentially limited by the simplicity of the hypotheses used. We have neither experimented with alternative zero-shot hypotheses nor tested few-shot classification, which could potentially improve the results -- especially on political leaning. Instead, we have focused on supervised learning.

Sixth, we have not carried out extensive testing of our newly trained models. This is required to ensure they have not learned unintended arbitrary features such as specific media outlet characteristics or stylistic patterns (as shown by \cite{baly-etal-2020-detect}, rather than political leaning itself. Evaluation on a set of handcrafted examples or real texts written in the future could potentially reveal some flaws.

\section{Future work}

Our research reveals several promising directions for future work.

First, a deeper manual analysis of our models' incorrect predictions could yield valuable insights into potential error patterns. Such investigation would help identify whether certain political topics, writing styles, or nuanced political positions consistently challenge the classifiers, informing targeted improvements to the models.

Building on our current three-class political leaning system, future work could implement a more granular political spectrum with five or more levels. This would capture the continuum from far-left to far-right more precisely, better reflecting the reality of political discourse. Additionally, introducing a ``mixed'' class would acknowledge texts that exhibit support for multiple political positions simultaneously. Another promising approach could be to encode the center class as semantically positioned between left and right, perhaps using a continuous numerical scale rather than discrete categories.

Instead of maintaining separate classifiers for politicalness and political leaning, a unified approach could be explored by adding a fourth ``non-political'' class to the political leaning model. This would simplify the classification pipeline and potentially allow the model to better understand the relationship between political and non-political content.

With access to more powerful computing resources than those available for this research (details in appendix \ref{sec:appendix_hardware}), comprehensive hyperparameter optimization could be conducted on larger models like DeBERTa V3 large. Our research suggests that larger models can perform better on out-of-distribution data, and optimizing their hyperparameters specifically for political text classification could yield significant improvements.

To address data limitations, creating additional synthetic political texts using generative LLMs could help expand and balance our datasets. Similarly, leveraging LLMs to validate or refine existing labels might improve the annotation quality, particularly for edge cases.

Another important extension would be developing multilingual models capable of classifying political content across languages. This would be particularly valuable as political discourse increasingly occurs in global contexts and would require assembling diverse multilingual political datasets.

Finally, improving the interpretability and explainability of our models represents a critical direction for future work, as appealed to by \cite{doan-gulla-2022-survey, nemeth2023scoping}. Developing methods to highlight which parts of a text most strongly influenced the model's prediction would make these tools more transparent and useful for both research and applications.

\section{Conclusion}

In this work, we have addressed the challenge of automatically classifying text according to political leaning and politicalness. We have collected and selected 12 datasets with annotations for political leaning. For politicalness, we have combined 18 datasets and unified their diverse labels for the binary classification. Our comprehensive approach involved evaluating existing models, and training new ones with enhanced generalization capabilities. Extensive benchmarking through leave-one-in and leave-one-out methodologies has provided valuable insights into the transferability of these models across different text domains. We have demonstrated that the existing NLI classifier Political DEBATE performs remarkably well on the politicalness task, while our newly trained models based on POLITICS and DeBERTa large establish a new state-of-the-art performance for political leaning classification.

An unpleasant, but important, observation is that all tested models (both existing and our newly trained) perform consistently worse on datasets they have not been trained on. These findings are in line with our proposed hypothesis as well as prior literature. We have made steps to address and overcome this phenomenon, which have led to improved performance on out-of-distribution texts. The fall of accuracy is, however, still present. We conclude that it truly is a hard task to teach a model to classify texts by political leaning universally -- on a wider range of topics and on a different character of text than the ones present in the training set.

The implications of this research extend beyond academic interest to practical applications in media analysis, automated content curation, and promoting balanced information consumption. Our findings highlight the importance of diversity in training data when developing models for political text classification, as well as the need for careful evaluation on examples from unseen distributions. Our work provides both valuable tools and methodological insights for researchers and practitioners analyzing politically biased content in a polarized information landscape.

\section*{Acknowledgement}
This work was funded by the European Union NextGenerationEU through the Recovery and Resilience Plan for Slovakia under the project No. 09I03-03-V03-00020.

\bibliography{anthology, custom}

\appendix

\section{Dataset intersection methodology details}\label{sec:appendix_methodology_dataset_intersection}

There is a total of nine possible combinations of the row pair's titles and bodies presence or absence. Each one requires its own comparison approach, as shown in table \ref{fig:dataset_intersection_comparisons}.

\begin{table*}[!ht]
    \centering
    \small
    \begin{tabular}{ll|lll}
    \toprule
    & & \multicolumn{3}{c}{\textbf{1st dataset's row}} \\ 
    & & \textbf{only has a title} & \textbf{only has a body} & \textbf{has both} \\ \midrule
    \multirow{3}{*}{\textbf{2nd dataset's row}} & \textbf{only has a title} & titles are equal & body contains title & titles are equal \\ 
    & \textbf{only has a body} & body contains title & 2nd body contains 1st slice & 2nd body contains 1st slice \\ 
    & \textbf{has both} & titles are equal & 2nd body contains 1st slice & titles are equal \\ 
    \bottomrule
    \end{tabular}
    \caption{The selected comparisons based on the field presence or absence for the dataset intersection.}
    \label{fig:dataset_intersection_comparisons}
\end{table*}

\section{Complete resulting tables}

\subsection{Dataset intersection}\label{sec:appendix_dataset_intersection}

Table \ref{fig:dataset_intersection}.

\begin{table*}[ht]
    \centering
    \small
    \begin{tabular}{lrrrrrrrrrrrr}
    \toprule
    ~ & \rotatebox[origin=l]{90}{Article bias prediction} & \rotatebox[origin=l]{90}{BIGNEWSBLN} & \rotatebox[origin=l]{90}{CommonCrawl news articles} & \rotatebox[origin=l]{90}{Dem., rep. party platform topics} & \rotatebox[origin=l]{90}{GPT-4 political bias} & \rotatebox[origin=l]{90}{GPT-4 political ideologies} & \rotatebox[origin=l]{90}{Media political stance} & \rotatebox[origin=l]{90}{Political podcasts} & \rotatebox[origin=l]{90}{Political tweets} & \rotatebox[origin=l]{90}{Qbias} & \rotatebox[origin=l]{90}{Webis bias flipper 18} & \rotatebox[origin=l]{90}{Webis news bias 20} \\ 
    \midrule
    Article bias prediction & \textcolor{gray}{--} & 1 & 2 & \textcolor{gray}{0} & \textcolor{gray}{0} & \textcolor{gray}{0} & 2 & \textcolor{gray}{0} & \textcolor{gray}{0} & \textcolor{gray}{0} & 7 & 10 \\ 
    BIGNEWSBLN & \textcolor{gray}{0} & \textcolor{gray}{--} & 1 & \textcolor{gray}{0} & \textcolor{gray}{0} & \textcolor{gray}{0} & 1 & \textcolor{gray}{0} & \textcolor{gray}{0} & \textcolor{gray}{0} & \textcolor{gray}{0} & \textcolor{gray}{0} \\ 
    CommonCrawl news articles & 1 & 1 & \textcolor{gray}{--} & \textcolor{gray}{0} & \textcolor{gray}{0} & \textcolor{gray}{0} & 2 & \textcolor{gray}{0} & \textcolor{gray}{0} & \textcolor{gray}{0} & \textcolor{gray}{0} & \textcolor{gray}{0} \\ 
    Dem., rep. party platform topics & \textcolor{gray}{0} & \textcolor{gray}{0} & \textcolor{gray}{0} & \textcolor{gray}{--} & \textcolor{gray}{0} & \textcolor{gray}{0} & \textcolor{gray}{0} & \textcolor{gray}{0} & \textcolor{gray}{0} & \textcolor{gray}{0} & \textcolor{gray}{0} & \textcolor{gray}{0} \\ 
    GPT-4 political bias & \textcolor{gray}{0} & \textcolor{gray}{0} & \textcolor{gray}{0} & \textcolor{gray}{0} & \textcolor{gray}{--} & \textcolor{gray}{0} & \textcolor{gray}{0} & \textcolor{gray}{0} & \textcolor{gray}{0} & \textcolor{gray}{0} & \textcolor{gray}{0} & \textcolor{gray}{0} \\ 
    GPT-4 political ideologies & \textcolor{gray}{0} & \textcolor{gray}{0} & \textcolor{gray}{0} & \textcolor{gray}{0} & \textcolor{gray}{0} & \textcolor{gray}{--} & \textcolor{gray}{0} & \textcolor{gray}{0} & \textcolor{gray}{0} & \textcolor{gray}{0} & \textcolor{gray}{0} & \textcolor{gray}{0} \\ 
    Media political stance & 1 & 1 & 2 & \textcolor{gray}{0} & \textcolor{gray}{0} & \textcolor{gray}{0} & \textcolor{gray}{--} & \textcolor{gray}{0} & \textcolor{gray}{0} & \textcolor{gray}{0} & \textcolor{gray}{0} & \textcolor{gray}{0} \\ 
    Political podcasts & \textcolor{gray}{0} & \textcolor{gray}{0} & \textcolor{gray}{0} & \textcolor{gray}{0} & \textcolor{gray}{0} & \textcolor{gray}{0} & \textcolor{gray}{0} & \textcolor{gray}{--} & \textcolor{gray}{0} & \textcolor{gray}{0} & \textcolor{gray}{0} & \textcolor{gray}{0} \\ 
    Political tweets & \textcolor{gray}{0} & \textcolor{gray}{0} & \textcolor{gray}{0} & \textcolor{gray}{0} & \textcolor{gray}{0} & \textcolor{gray}{0} & \textcolor{gray}{0} & \textcolor{gray}{0} & \textcolor{gray}{--} & \textcolor{gray}{0} & \textcolor{gray}{0} & \textcolor{gray}{0} \\ 
    Qbias & \textcolor{gray}{0} & \textcolor{gray}{0} & \textcolor{gray}{0} & \textcolor{gray}{0} & \textcolor{gray}{0} & \textcolor{gray}{0} & \textcolor{gray}{0} & \textcolor{gray}{0} & \textcolor{gray}{0} & \textcolor{gray}{--} & \textcolor{gray}{0} & 1 \\ 
    Webis bias flipper 18 & 42 & 1 & 4 & \textcolor{gray}{0} & \textcolor{gray}{0} & \textcolor{gray}{0} & 3 & \textcolor{gray}{0} & 1 & 1 & \textcolor{gray}{--} & 80 \\ 
    Webis news bias 20 & 50 & 1 & 3 & \textcolor{gray}{0} & \textcolor{gray}{0} & \textcolor{gray}{0} & 3 & \textcolor{gray}{0} & 1 & 2 & 66 & \textcolor{gray}{--} \\ 
    \bottomrule
    \end{tabular}
    \vfill
    \vspace{0.5cm}
    \begin{tabular}{lrrrrrrrrrrrrrrrrrr}
    \toprule
    ~ & \rotatebox[origin=l]{90}{Amazon reviews 2023} & \rotatebox[origin=l]{90}{Dialogsum} & \rotatebox[origin=l]{90}{Free news} & \rotatebox[origin=l]{90}{Goodreads book genres} & \rotatebox[origin=l]{90}{IMDB} & \rotatebox[origin=l]{90}{IMDB movie genres} & \rotatebox[origin=l]{90}{Medium post titles} & \rotatebox[origin=l]{90}{News category} & \rotatebox[origin=l]{90}{PENS} & \rotatebox[origin=l]{90}{PoliBERTweet} & \rotatebox[origin=l]{90}{Political or not} & \rotatebox[origin=l]{90}{Recipes} & \rotatebox[origin=l]{90}{Reddit comments} & \rotatebox[origin=l]{90}{Reddit submissions} & \rotatebox[origin=l]{90}{Rotten tomatoes} & \rotatebox[origin=l]{90}{Textbooks} & \rotatebox[origin=l]{90}{Tweet topic multi} & \rotatebox[origin=l]{90}{Yelp review full} \\
    \midrule
    Amazon reviews & \textcolor{gray}{--} & \textcolor{gray}{0} & \textcolor{gray}{0} & \textcolor{gray}{0} & 1 & \textcolor{gray}{0} & 1 & \textcolor{gray}{0} & \textcolor{gray}{0} & \textcolor{gray}{0} & \textcolor{gray}{0} & 1 & 1 & 1 & \textcolor{gray}{0} & \textcolor{gray}{0} & \textcolor{gray}{0} & 1 \\
    Dialogsum & 2 & \textcolor{gray}{--} & \textcolor{gray}{0} & \textcolor{gray}{0} & \textcolor{gray}{0} & \textcolor{gray}{0} & 2 & \textcolor{gray}{0} & \textcolor{gray}{0} & \textcolor{gray}{0} & \textcolor{gray}{0} & \textcolor{gray}{0} & \textcolor{gray}{0} & \textcolor{gray}{0} & \textcolor{gray}{0} & \textcolor{gray}{0} & \textcolor{gray}{0} & \textcolor{gray}{0} \\
    Free news & \textcolor{gray}{0} & \textcolor{gray}{0} & \textcolor{gray}{--} & \textcolor{gray}{0} & \textcolor{gray}{0} & \textcolor{gray}{0} & \textcolor{gray}{0} & \textcolor{gray}{0} & \textcolor{gray}{0} & \textcolor{gray}{0} & \textcolor{gray}{0} & \textcolor{gray}{0} & \textcolor{gray}{0} & \textcolor{gray}{0} & \textcolor{gray}{0} & \textcolor{gray}{0} & \textcolor{gray}{0} & \textcolor{gray}{0} \\
    Goodreads book genres & 1 & \textcolor{gray}{0} & \textcolor{gray}{0} & \textcolor{gray}{--} & \textcolor{gray}{0} & \textcolor{gray}{0} & 5 & \textcolor{gray}{0} & \textcolor{gray}{0} & \textcolor{gray}{0} & \textcolor{gray}{0} & \textcolor{gray}{0} & \textcolor{gray}{0} & \textcolor{gray}{0} & \textcolor{gray}{0} & \textcolor{gray}{0} & \textcolor{gray}{0} & \textcolor{gray}{0} \\
    IMDB & 1 & \textcolor{gray}{0} & \textcolor{gray}{0} & \textcolor{gray}{0} & \textcolor{gray}{--} & \textcolor{gray}{0} & 4 & \textcolor{gray}{0} & \textcolor{gray}{0} & \textcolor{gray}{0} & \textcolor{gray}{0} & \textcolor{gray}{0} & \textcolor{gray}{0} & \textcolor{gray}{0} & \textcolor{gray}{0} & \textcolor{gray}{0} & \textcolor{gray}{0} & \textcolor{gray}{0} \\
    IMDB movie genres & \textcolor{gray}{0} & \textcolor{gray}{0} & \textcolor{gray}{0} & \textcolor{gray}{0} & \textcolor{gray}{0} & \textcolor{gray}{--} & 3 & \textcolor{gray}{0} & \textcolor{gray}{0} & \textcolor{gray}{0} & \textcolor{gray}{0} & \textcolor{gray}{0} & \textcolor{gray}{0} & \textcolor{gray}{0} & \textcolor{gray}{0} & \textcolor{gray}{0} & \textcolor{gray}{0} & \textcolor{gray}{0} \\
    Medium post titles & 1 & \textcolor{gray}{0} & \textcolor{gray}{0} & 1 & 8 & 7 & \textcolor{gray}{--} & \textcolor{gray}{0} & \textcolor{gray}{0} & \textcolor{gray}{0} & \textcolor{gray}{0} & \textcolor{gray}{0} & 1 & \textcolor{gray}{0} & \textcolor{gray}{0} & \textcolor{gray}{0} & \textcolor{gray}{0} & \textcolor{gray}{0} \\
    News category & \textcolor{gray}{0} & \textcolor{gray}{0} & \textcolor{gray}{0} & \textcolor{gray}{0} & \textcolor{gray}{0} & \textcolor{gray}{0} & \textcolor{gray}{0} & \textcolor{gray}{--} & \textcolor{gray}{0} & \textcolor{gray}{0} & \textcolor{gray}{0} & \textcolor{gray}{0} & \textcolor{gray}{0} & \textcolor{gray}{0} & \textcolor{gray}{0} & \textcolor{gray}{0} & \textcolor{gray}{0} & \textcolor{gray}{0} \\
    PENS & \textcolor{gray}{0} & \textcolor{gray}{0} & \textcolor{gray}{0} & \textcolor{gray}{0} & \textcolor{gray}{0} & \textcolor{gray}{0} & \textcolor{gray}{0} & \textcolor{gray}{0} & \textcolor{gray}{--} & \textcolor{gray}{0} & \textcolor{gray}{0} & \textcolor{gray}{0} & \textcolor{gray}{0} & \textcolor{gray}{0} & \textcolor{gray}{0} & \textcolor{gray}{0} & \textcolor{gray}{0} & \textcolor{gray}{0} \\
    PoliBERTweet & \textcolor{gray}{0} & \textcolor{gray}{0} & \textcolor{gray}{0} & \textcolor{gray}{0} & \textcolor{gray}{0} & \textcolor{gray}{0} & \textcolor{gray}{0} & 1 & \textcolor{gray}{0} & \textcolor{gray}{--} & \textcolor{gray}{0} & \textcolor{gray}{0} & 1 & \textcolor{gray}{0} & \textcolor{gray}{0} & \textcolor{gray}{0} & \textcolor{gray}{0} & \textcolor{gray}{0} \\
    Political or not & 3 & \textcolor{gray}{0} & \textcolor{gray}{0} & \textcolor{gray}{0} & \textcolor{gray}{0} & \textcolor{gray}{0} & 2 & 3 & \textcolor{gray}{0} & 1 & \textcolor{gray}{--} & \textcolor{gray}{0} & \textcolor{gray}{0} & \textcolor{gray}{0} & \textcolor{gray}{0} & \textcolor{gray}{0} & \textcolor{gray}{0} & \textcolor{gray}{0} \\
    Recipes & \textcolor{gray}{0} & \textcolor{gray}{0} & \textcolor{gray}{0} & \textcolor{gray}{0} & \textcolor{gray}{0} & \textcolor{gray}{0} & \textcolor{gray}{0} & \textcolor{gray}{0} & \textcolor{gray}{0} & \textcolor{gray}{0} & \textcolor{gray}{0} & \textcolor{gray}{--} & \textcolor{gray}{0} & \textcolor{gray}{0} & \textcolor{gray}{0} & \textcolor{gray}{0} & \textcolor{gray}{0} & \textcolor{gray}{0} \\
    Reddit comments & \textcolor{gray}{0} & \textcolor{gray}{0} & \textcolor{gray}{0} & \textcolor{gray}{0} & \textcolor{gray}{0} & \textcolor{gray}{0} & \textcolor{gray}{0} & \textcolor{gray}{0} & \textcolor{gray}{0} & \textcolor{gray}{0} & \textcolor{gray}{0} & \textcolor{gray}{0} & \textcolor{gray}{--} & 1 & \textcolor{gray}{0} & \textcolor{gray}{0} & \textcolor{gray}{0} & 1 \\
    Reddit submissions & 1 & \textcolor{gray}{0} & \textcolor{gray}{0} & \textcolor{gray}{0} & \textcolor{gray}{0} & \textcolor{gray}{0} & \textcolor{gray}{0} & \textcolor{gray}{0} & \textcolor{gray}{0} & \textcolor{gray}{0} & \textcolor{gray}{0} & \textcolor{gray}{0} & 3 & \textcolor{gray}{--} & \textcolor{gray}{0} & \textcolor{gray}{0} & \textcolor{gray}{0} & \textcolor{gray}{0} \\
    Rotten tomatoes & 1 & \textcolor{gray}{0} & \textcolor{gray}{0} & \textcolor{gray}{0} & \textcolor{gray}{0} & \textcolor{gray}{0} & \textcolor{gray}{0} & 1 & \textcolor{gray}{0} & \textcolor{gray}{0} & \textcolor{gray}{0} & \textcolor{gray}{0} & 3 & \textcolor{gray}{0} & \textcolor{gray}{--} & \textcolor{gray}{0} & \textcolor{gray}{0} & \textcolor{gray}{0} \\
    Textbooks & \textcolor{gray}{0} & \textcolor{gray}{0} & \textcolor{gray}{0} & \textcolor{gray}{0} & \textcolor{gray}{0} & \textcolor{gray}{0} & \textcolor{gray}{0} & \textcolor{gray}{0} & \textcolor{gray}{0} & \textcolor{gray}{0} & \textcolor{gray}{0} & \textcolor{gray}{0} & \textcolor{gray}{0} & \textcolor{gray}{0} & \textcolor{gray}{0} & \textcolor{gray}{--} & \textcolor{gray}{0} & \textcolor{gray}{0} \\
    Tweet topic multi & \textcolor{gray}{0} & \textcolor{gray}{0} & \textcolor{gray}{0} & \textcolor{gray}{0} & \textcolor{gray}{0} & \textcolor{gray}{0} & \textcolor{gray}{0} & \textcolor{gray}{0} & \textcolor{gray}{0} & \textcolor{gray}{0} & \textcolor{gray}{0} & \textcolor{gray}{0} & 2 & \textcolor{gray}{0} & \textcolor{gray}{0} & \textcolor{gray}{0} & \textcolor{gray}{--} & \textcolor{gray}{0} \\
    Yelp review full & 1 & \textcolor{gray}{0} & \textcolor{gray}{0} & \textcolor{gray}{0} & \textcolor{gray}{0} & \textcolor{gray}{0} & \textcolor{gray}{0} & \textcolor{gray}{0} & \textcolor{gray}{0} & \textcolor{gray}{0} & \textcolor{gray}{0} & \textcolor{gray}{0} & 2 & \textcolor{gray}{0} & \textcolor{gray}{0} & \textcolor{gray}{0} & \textcolor{gray}{0} & \textcolor{gray}{--} \\
    \bottomrule
    \end{tabular}
    \caption{The resulting intersections between every pair of the political leaning (the first table) and politicalness (the second table) datasets. The numbers are percentages and they mean the portion of the intersection size over the total number of examples in the dataset in the corresponding row. For example in the political leaning datasets, Article bias prediction has an intersection of 740 instances with CommonCrawl news articles, which is 2~\% of Article bias prediction. In the CommonCrawl news articles row, the same intersection size (740) amounts to just 1~\% of that dataset.}
    \label{fig:dataset_intersection}
\end{table*}

\subsection{Existing politicalness models}\label{sec:appendix_existing_evaluation_politicalness}

Table \ref{fig:models_eval_existing_politicalness2}.

\begin{table*}[ht]
    \centering
    \small
    \begin{tabular}{lrrr}
    \toprule
    ~ & Classifier main subject politics & Political DEBATE large & Topic politics \\ 
    \midrule
    Amazon reviews & 99.8 & 99.4 & 100 \\ 
    Dialogsum & 99.6 & 98.8 & 99.9 \\ 
    Free news & 73.8 & 82.8 & 67.4 \\ 
    Goodreads book genres & 99.3 & 99.2 & 100 \\ 
    IMDB & 99.9 & 99.7 & 100 \\ 
    IMDB movie genres & 98.6 & 97.8 & 99.7 \\ 
    Medium post titles & 62.9 & 93.7 & 81.9 \\ 
    News category & 68.8 & 95.3 & 80.3 \\ 
    PENS & 75.4 & 97.8 & 87.9 \\ 
    PoliBERTweet & 56 & 85.9 & 81.7 \\ 
    Political or not & 73.5 & 99.8 & 83.4 \\ 
    Recipes & 100 & 100 & 100 \\ 
    Reddit comments & 98.2 & 95.2 & 99.3 \\ 
    Reddit submissions & 99.3 & 98.5 & 99.7 \\ 
    Rotten tomatoes & 99.8 & 96.2 & 99.6 \\ 
    Textbooks & 58 & 95.6 & 54.5 \\ 
    Tweet topic multi & 99.5 & 98.6 & 99.8 \\ 
    Yelp review full & 99.8 & 99.8 & 100 \\ 
    Article bias prediction & 69.6 & 93.4 & 76.7 \\ 
    BIGNEWSBLN & 69.6 & 84.3 & 65 \\ 
    CommonCrawl news articles & 71.1 & 98.8 & 70.5 \\ 
    Dem., rep. party platform topics & 70.1 & 98.2 & 81.4 \\ 
    GPT-4 political bias & 66.5 & 99.8 & 48.9 \\ 
    GPT-4 political ideologies & 18.3 & 98.6 & 86.9 \\ 
    Media political stance & 60.3 & 81.4 & 57.4 \\ 
    Political podcasts & 10.2 & 89.7 & 41.5 \\ 
    Political tweets & 54.3 & 85.2 & 56.4 \\ 
    Qbias & 58.5 & 90.7 & 68 \\ 
    Webis bias flipper 18 & 75.2 & 92.2 & 81.8 \\ 
    Webis news bias 20 & 71.9 & 94.6 & 78.2 \\ 
    \bottomrule
    \end{tabular}
    \caption{Resulting $F_1$ scores after evaluating the existing politicalness models on all available (both politicalness and political leaning) datasets, sampled to 1~000 examples each. The class distributions of the datasets need to be taken into account when interpreting these metrics, as explained in section \ref{sec:methodology_existing_evaluation_politicalness}. For a simplified, balanced and clear comparison, refer to section \ref{sec:results_existing_evaluation_politicalness}.}
    \label{fig:models_eval_existing_politicalness2}
\end{table*}

\subsection{Politicalness leave-one-out dataset benchmark}\label{sec:appendix_leave_one_out_politicalness}

Table \ref{fig:models_eval_out_politicalness_complete}.

\begin{table*}[ht]
    \centering
    \small
    \begin{tabular}{lrrrrrrrr}
    \toprule
    \textsc{the left-out dataset} $\downarrow$ & \multicolumn{2}{c}{BERT} & \multicolumn{2}{c}{RoBERTa} & \multicolumn{2}{c}{DeBERTa} & \multicolumn{2}{c}{POLITICS} \\ 
    \textsc{test set distribution} $\rightarrow$ & trained & left-out & trained & left-out & trained & left-out & trained & left-out \\ 
    \midrule
    Amazon reviews & 97.6 & 99.9 & 97.9 & 100 & 97.9 & 99.9 & 97.8 & 100 \\ 
    Dialogsum & 97.6 & 99 & 97.9 & 98.7 & 97.9 & 98.1 & 97.6 & 98.5 \\ 
    Free news & 97.5 & 96.9 & 97.9 & 97.2 & 97.9 & 97.8 & 98 & 95.7 \\ 
    Goodreads book genres & 97.3 & 98.6 & 97.6 & 98.7 & 97 & 99.3 & 97.8 & 99.2 \\ 
    IMDB & 97.4 & 98.3 & 97.3 & 94.6 & 97.8 & 96.6 & 97.9 & 98.9 \\ 
    IMDB movie genres & 97.5 & 99.6 & 97.2 & 98.9 & 97.9 & 99.5 & 97.8 & 99.8 \\ 
    Medium post titles & 98 & 69.1 & 98.4 & 65 & 98.4 & 67.9 & 98.4 & 71.7 \\ 
    News category & 97.7 & 99.2 & 97.7 & 96.8 & 97.8 & 98.9 & 97.4 & 92.3 \\ 
    PENS & 97.5 & 100 & 98 & 100 & 98 & 100 & 97.7 & 100 \\ 
    PoliBERTweet & 97.7 & 98.8 & 97.2 & 99.8 & 97.7 & 94.6 & 97.7 & 96.3 \\ 
    Political or not & 97.6 & 99.8 & 97.4 & 99.8 & 97.8 & 99.8 & 97.8 & 99.7 \\ 
    Recipes & 97.5 & 99 & 97.1 & 96.4 & 96.8 & 94.3 & 97.9 & 99 \\ 
    Reddit comments & 97.6 & 92.7 & 97.9 & 85.5 & 98.1 & 91.4 & 98.1 & 90.8 \\ 
    Reddit submissions & 98.1 & 88.8 & 97.6 & 87.1 & 98.3 & 90.8 & 98.4 & 88.6 \\ 
    Rotten tomatoes & 97.6 & 86.8 & 97.8 & 92.3 & 98.1 & 92.1 & 97.7 & 86.6 \\ 
    Textbooks & 97.7 & 91.3 & 97.8 & 90.8 & 97.9 & 90.4 & 98.3 & 78 \\ 
    Tweet topic multi & 97.7 & 89.9 & 50.7 & 33.3 & 97.5 & 92.9 & 97.9 & 84.7 \\ 
    Yelp review full & 97.6 & 79.4 & 97.8 & 79.4 & 98 & 77.3 & 97.8 & 75.8 \\ 
    Article bias prediction & 97.6 & 89.8 & 97.7 & 87.2 & 98 & 94.1 & 97.6 & 87.5 \\ 
    BIGNEWSBLN & 97.6 & 51.3 & 97.9 & 62 & 97.3 & 76.6 & 97.7 & 73.9 \\ 
    CommonCrawl news articles & 97.6 & 97.7 & 97.6 & 100 & 98 & 97.7 & 97.9 & 96.1 \\ 
    Dem., rep. party platform topics & 97.6 & 100 & 95 & 98.6 & 95.8 & 100 & 97.8 & 100 \\ 
    GPT-4 political bias & 97.6 & 98 & 97.9 & 96.7 & 97.8 & 96.8 & 98 & 96.4 \\ 
    GPT-4 political ideologies & 97.5 & 99 & 97.7 & 99 & 97.9 & 99.6 & 97.9 & 99.6 \\ 
    Media political stance & 97.6 & 99.2 & 97.1 & 98.9 & 97.4 & 99.7 & 97.9 & 99.3 \\ 
    Political podcasts & 97.7 & 93.6 & 97.9 & 93.3 & 98.1 & 95.3 & 97.9 & 90.8 \\ 
    Political tweets & 97.5 & 96.7 & 97.1 & 98.1 & 97.1 & 95.1 & 97.6 & 91.9 \\ 
    Qbias & 97.6 & 99.9 & 97.8 & 99.9 & 97.8 & 100 & 97.5 & 99.9 \\ 
    \bottomrule
    \end{tabular}
    \caption{Resulting F1 scores on politicalness classification after evaluating the base versions of the BERT (cased), RoBERTa, DeBERTa V3 and POLITICS models (test set size: 15~\%; train set size: 1~000 from each dataset) fine-tuned on all available datasets (28) except for one (the left-out) -- as described in section \ref{sec:methodology_benchmark_leave_one_out}. The values in the trained distribution columns are the average score on all the datasets the model has been trained on. The left-out columns are the score on the one left-out dataset. For averages of these scores, refer to section \ref{sec:results_dataset_benchmark_out_politicalness}.}
    \label{fig:models_eval_out_politicalness_complete}
\end{table*}

\section{Hardware and durations}\label{sec:appendix_hardware}

The dataset intersections have been measured on Intel Core i7-3770, 16~GB RAM. It took 50~hours.

On a machine with NVIDIA RTX~2060, AMD Ryzen~7~4800H and 16~GB RAM, the existing models have been evaluated (about 5~minutes per model on a single dataset) and the prototype models for dataset benchmarks without optimizing hyperparameters have been fine-tuned (about 15~minutes per model for leave-one-in, 35~minutes per model for leave-one-out).

The hyperparameter search for POLITICS has been ran on NVIDIA RTX~4000~Ada, 5~vCPUs, 47~GB RAM and took 70~minutes per trial.

The tuning of hyperparameters for DeBERTa~V3 large has been done on NVIDIA RTX~4070~Ti, 8~vCPUs, 30~GB RAM and one epoch took approximately 1~hour.

\end{document}